\documentclass{article}

\usepackage{cite}

\usepackage[final,nonatbib]{nips_2018}

\usepackage[utf8]{inputenc} % allow utf-8 input
\usepackage[T1]{fontenc}    % use 8-bit T1 fonts
\usepackage{hyperref}       % hyperlinks
\usepackage{url}            % simple URL typesetting
\usepackage{booktabs}       % professional-quality tables
\usepackage{amsfonts}       % blackboard math symbols
\usepackage{nicefrac}       % compact symbols for 1/2, etc.
\usepackage{microtype}      % microtypography

\usepackage{subcaption}
\usepackage{algorithm}
\usepackage[noend]{algpseudocode}
\usepackage{enumitem}

\usepackage{mathtools}
\usepackage[utf8]{inputenc}
\usepackage[english]{babel}
\usepackage{amssymb,amsmath,amsthm}

\newtheorem{theorem}{Theorem}
\newtheorem{lemma}{Lemma}

\DeclareMathOperator*{\argmin}{arg\,min}
\DeclareMathOperator*{\argmax}{arg\,max}

\newcommand{\RR}[0]{\mathbb{R}}

\newcommand{\rbar}{\overline{r}}
\newcommand{\vbar}{\overline{v}}
\newcommand{\vtheta}{\boldsymbol{\theta}}
\newcommand{\Rc}{\mathcal{R}}
\newcommand{\Jbad}{J_{\mathrm{bad}}}
\newcommand{\Rbad}{\mathcal{R}_{\mathrm{bad}}}

\newcommand{\fpoly}{f_{\mathrm{poly}}}
\newcommand{\gpoly}{g_{\mathrm{poly}}}

\def\l({\left(}
\def\r){\right)}
\def\bl({\Big(}
\def\br){\Big)}
\def\beq{\begin{equation}}
\def\eeq{\end{equation}}

\usepackage{tikz,pgfplots}
\def\x{{\mathbf x}}
\def\y{{\mathbf y}}

\def\z{{\mathbf z}}

\def\m{{\mathbf m}}

\def\RR{{\mathbb R}}

\def\I{{ \mathbf I }}

\def\bone{{\boldsymbol{1}}}

\def\PP{{ \mathbb P }}
\def\EE{{ \mathbb E }}
\def\ZZ{{ \mathbb Z }}

\def\xtil{{ \tilde{\mathbf x} }}

\newcommand{\Normal}{\mathcal{N}}
\newcommand{\Hk}{\mathcal{H}_{k}}

\newcommand{\vk}{\mathbf{k}}
\newcommand{\vK}{\mathbf{K}}

\newcommand{\vy}{\mathbf{y}}
\newcommand{\vdelta}{\boldsymbol{\delta}}
\newcommand{\alg}{\textsc{StableOpt}}
\newcommand{\ucb}{\mathrm{ucb}}
\newcommand{\lcb}{\mathrm{lcb}}

\newcommand{\gpucb}{\textsc{GP-UCB}}
\newcommand{\mmgpucb}{\textsc{MaxiMin-GP-UCB}}
\newcommand{\sgpucb}{\textsc{Stable-GP-UCB}}
\newcommand{\srandom}{\textsc{Stable-GP-Random}}

\newcommand{\Fc}{\mathcal{F}}

\newcommand{\Dbar}{\overline{D}}
\newcommand{\kSE}{k_{\text{SE}}}
\newcommand{\kMat}{k_{\text{Mat\'ern}}}

\newcommand{\R}{\textbf{R}}
\newcommand{\vu}{\textbf{u}}
\newcommand{\vr}{\boldsymbol{r}}

\title{Adversarially Robust Optimization \\ with Gaussian Processes}

\author{
  Ilija Bogunovic \\
  LIONS, EPFL \\
  \texttt{ilija.bogunovic@epfl.ch} \\
   \And
   Jonathan Scarlett \\
   National University of Singapore \\
   \texttt{scarlett@comp.nus.edu.sg} \\
   \AND
   Stefanie Jegelka \\
   MIT CSAIL \\
   \texttt{stefje@mit.edu} \\
   \And
   Volkan Cevher \\
   LIONS, EPFL \\
   \texttt{volkan.cevher@epfl.ch} 
}

\begin{document}

\maketitle

%!TEX root = main.tex
\begin{abstract}
    In this paper, we consider the problem of Gaussian process (GP) optimization with an added robustness requirement: The returned point may be perturbed by an adversary, and we require the function value to remain as high as possible even after this perturbation. This problem is motivated by settings in which the underlying functions during optimization and implementation stages are different, % (e.g., due to time variations), 
    or when one is interested in finding an entire region of good inputs rather than only a single point.  We show that standard GP optimization algorithms do not exhibit the desired robustness properties, and provide a novel confidence-bound based algorithm \alg~for this purpose.  We rigorously establish the required number of samples for \alg~to find a near-optimal point, and we complement this guarantee with an algorithm-independent lower bound.  We experimentally demonstrate several potential applications of interest using real-world data sets, and we show that \alg~consistently succeeds in finding a stable maximizer where several baseline methods fail.
\end{abstract}
%!TEX root = main.tex
\section{Introduction}

Gaussian processes (GP) provide a powerful means for sequentially optimizing a black-box function $f$ that is costly to evaluate and for which noisy point evaluations are available.  Since its introduction, this approach has successfully been applied to numerous applications, including robotics \cite{lizotte2007automatic}, hyperparameter tuning \cite{snoek2012practical}, recommender systems \cite{vanchinathan2014explore}, environmental monitoring \cite{srinivas2009gaussian}, and more. 

In many such applications, one is faced with various forms of uncertainty that are not accounted for by standard algorithms.  In robotics, the optimization is often performed via simulations, creating a mismatch between the assumed function and the true one; in hyperparameter tuning, the function is typically similarly mismatched due to limited training data; in recommendation systems and several other applications, the underlying function is inherently time-varying, so the returned solution may become increasingly stale over time; the list goes on.

In this paper, we address these considerations by studying the GP optimization problem with an additional requirement of {\em adversarial robustness}: The returned point may be perturbed by an adversary, and we require the function value to remain as high as possible even after this perturbation.  This problem is of interest not only for attaining improved robustness to uncertainty, but also for settings where one seeks a region of good points rather than a single point, and for other related max-min optimization settings (see Section \ref{sec:variations} for further discussion).

\textbf{Related work.}
Numerous algorithms have been developed for GP optimization in recent years~\cite{srinivas2009gaussian, hennig2012entropy, hernandez2014predictive, bogunovic2016truncated, wang2017max, shekhar2017gaussian, ru2017fast}.  Beyond the standard setting, several important extensions have been considered, including batch sampling~\cite{desautels2014parallelizing, gonzalez2016batch, contal2013parallel},
contextual and time-varying settings~\cite{krause2011contextual, bogunovic2016time},
safety requirements~\cite{sui2015safe}, and
high dimensional settings~\cite{kandasamy2015high, wang2017batched, rolland2018high}, just to name a few.

Various forms of robustness in GP optimization have been considered previously.  A prominent example is that of outliers \cite{martinez2018practical}, in which certain function values are highly unreliable; however, this is a separate issue from that of the present paper, since in \cite{martinez2018practical} the returned point does not undergo any perturbation.  Another related recent work is \cite{beland2017bayes}, which assumes that the {\em sampled points} (rather than the returned one) are subject to uncertainty.  In addition to this difference, the uncertainty in \cite{beland2017bayes} is random rather than adversarial, which is complementary but distinct from our work. The same is true of a setting called {\em unscented Bayesian optimization} in \cite{nogueira2016unscented}.  Moreover, no theoretical results are given in \cite{beland2017bayes,nogueira2016unscented}.  In \cite{bogunovic2018robust}, a robust form of batch optimization is considered, but with yet another form of robustness, namely, some experiments in the batch may fail to produce an outcome.  Level-set estimation \cite{gotovos2013active,bogunovic2016truncated} is another approach to finding regions of good points rather than a single point.

Our problem formulation is also related to other works on non-convex robust optimization, particularly those of \textit{Bertsimas et al.}~\cite{bertsimas2010robust,bertsimas2010nonconvex}. In these works, a stable design $\x$ is sought that solves $\min_{\x \in D} \max_{\vdelta \in \mathcal{U}} f(\x + \vdelta)$. Here, $\vdelta$ resides in some uncertainty set $\mathcal{U}$, and represents the perturbation against which the design $\x$ needs to be protected. Related problems have also recently been considered in the context of adversarial training (e.g., \cite{sinha2017certifiable}).  Compared to these works, our work bears the crucial difference that the objective function is {\em unknown}, and we can only learn about it through noisy point evaluations (i.e. bandit feedback).  

Other works, such as \cite{chen2017robust,wilder2017equilibrium,staib2018distributionally,krause2008robust,bogunovic2017robust}, have considered robust optimization problems of the following form: For a given set of objectives  $\lbrace f_1, \dots, f_m \rbrace$ find $\x$ achieving $\max_{\x \in D} \min_{i = 1,\dotsc,m} f_i(\x)$.  We discuss variations of our algorithm for this type of formulation in Section \ref{sec:variations}.

\textbf{Contributions.} We introduce a variant of GP optimization in which the returned solution is required to exhibit stability/robustness to an adversarial perturbation.  We demonstrate the failures of standard algorithms, and introduce a new algorithm \alg~that overcomes these limitations.  We provide a novel theoretical analysis characterizing the number of samples required for \alg~to attain a near-optimal robust solution, and we complement this with an algorithm-independent lower bound.  We provide several variations of our max-min optimization framework and theory, including connections and comparisons to previous works.  Finally, we experimentally demonstrate a variety of potential applications of interest using real-world data sets, and we show that \alg~consistently succeeds in finding a stable maximizer where several baseline methods fail.

%!TEX root = main.tex

\section{Problem Setup} \label{sec:setup}
  
{\bf Model.} Let $f$ be an unknown reward function over a domain $D \subseteq \RR^p$ for some dimension $p$.   At time $t$, we query $f$ at a single point $\x_t \in D$ and observe a noisy sample $y_t = f(\x_t) + z_t$, where $z_t \sim \Normal(0,\sigma^2)$.  After $T$ rounds, a recommended point $\x^{(T)}$ is returned.  In contrast with the standard goal of making $f(\x^{(T)})$ as high as possible, we seek to find a point such that $f$ remains high even after an adversarial perturbation; a formal description is given below.
 
We assume that $D$ is endowed with a kernel function $k(\cdot, \cdot)$, and $f$ has a bounded norm in the corresponding Reproducing Kernel Hilbert Space (RKHS) $\Hk(D)$.  Specifically, we assume that $f \in \Fc_k(B)$, where
\begin{equation}
    \Fc_k(B) = \{ f \in \Hk(D) \,:\, \|f\|_{k} \leq B \}, \label{eq:func_class}
\end{equation}
and $\|f\|_{k}$ is the RKHS norm in  $\Hk(D)$.  It is well-known that this assumption permits the construction of confidence bounds via Gaussian process (GP) methods; see Lemma \ref{confidence_lemma} below for a precise statement.  We assume that the kernel is normalized to satisfy $k(\x,\x)=1$ for all $\x \in D$. Two commonly-considered kernels are squared exponential (SE) and Matérn:
\begin{gather}
k_{\text{SE}}(\x,\x') = \exp \left(- \dfrac{\|\x - \x'\|^2}{2l^2} \right), \\ k_{\text{Mat}}(\x,\x') = \frac{2^{1-\nu}}{ \Gamma(\nu) } \Big( \frac{\sqrt{2\nu} \|\x - \x'\| }{l} \Big) J_{\nu}\Big( \frac{\sqrt{2\nu} \|\x - \x'\| }{l} \Big),
\end{gather}
where $l$ denotes the length-scale, $\nu > 0$ is an additional parameter that dictates the smoothness,
and $J(\nu)$ and $\Gamma(\nu)$ denote the modified Bessel function and the gamma function, respectively \cite{rasmussen2006gaussian}.

Given a sequence of decisions $\lbrace \x_1, \cdots, \x_t \rbrace$ and their noisy observations $\lbrace y_1, \cdots, y_t \rbrace$, the posterior distribution under a $\mathrm{GP}(0, k(\x,\x'))$ prior is also Gaussian, with the following mean and variance:
\begin{gather}
\mu_{t}(\x) = \vk_t(\x)^T\big(\vK_t + \sigma^2 \mathbf{I} \big)^{-1} \vy_t,  \\ \sigma_{t}^2(\x) = k(\x,\x) - \vk_t(\x)^T \big(\vK_t + \sigma^2 \mathbf{I} \big)^{-1} \vk_t(\x),
\label{eq:mu_and_sigma_update}
\end{gather}
where $\vk_t(\x) = \big[k(\x_i,\x)\big]_{i=1}^t$, and $\vK_t = \big[k(\x_t,\x_{t'})\big]_{t,t'}$ is the kernel matrix.

\textbf{Optimization goal.} Let $d(\x,\x')$ be a function mapping $D \times D \to \RR$, and let $\epsilon$ be a constant known as the {\em stability parameter}.  For each point $\x \in D$, we define a set
\begin{equation}
    \Delta_{\epsilon}(\x) = \big\{ \x' - \x \,:\, \x' \in D \;\; \text{and}\;\; d(\x,\x') \le \epsilon \big\}.
\end{equation}
One can interpret this as the set of perturbations of $\x$ such that the newly obtained point $\x'$ is within a ``distance'' $\epsilon$ of $\x$.  While we refer to $d(\cdot, \cdot)$ as the distance function throughout the paper, we allow it to be a general function, and not necessarily a distance in the mathematical sense.   As we exemplify in Section \ref{sec:experiments}, the parameter $\epsilon$ might be naturally specified as part of the application, or might be better treated as a parameter that can be tuned for the purpose of the overall learning goal.

We define an {\em $\epsilon$-stable optimal input} to be any $\x^*_{\epsilon}$ satisfying
\begin{equation}
    \x^*_{\epsilon} \in \argmax_{\x \in D} \min_{\vdelta \in \Delta_{\epsilon}(\x)} f(\x + \vdelta). \label{eq:eps_stable_input}
\end{equation}
Our goal is to report a point $\x^{(T)}$ that is stable in the sense of having low {\em  $\epsilon$-regret}, defined as
\begin{equation} 
  \label{eq:eps_regret}
    r_{\epsilon}(\x) = \min_{\vdelta \in \Delta_{\epsilon}(\x^*_{\epsilon})} f(\x^*_{\epsilon} + \vdelta) - \min_{\vdelta \in \Delta_{\epsilon}(\x)} f(\x + \vdelta).
\end{equation}
Note that once $r_{\epsilon}(\x) \leq \eta$ for some accuracy value $\eta \ge 0$, it follows that 
\begin{equation}
  \label{eq:benchmark}
  \min_{\vdelta \in \Delta_{\epsilon}(\x)} f(\x + \vdelta) \geq \min_{\vdelta \in \Delta_{\epsilon}(\x^*_{\epsilon})} f(\x^*_{\epsilon}+ \vdelta) - \eta.  
\end{equation}
We assume that $d(\cdot, \cdot)$ and $\epsilon$ are known, i.e., they are specified as part of the optimization formulation.

\begin{figure}
          \centering
          	\begin{subfigure}{.32\textwidth}
              \centering
              \includegraphics[scale=0.35]{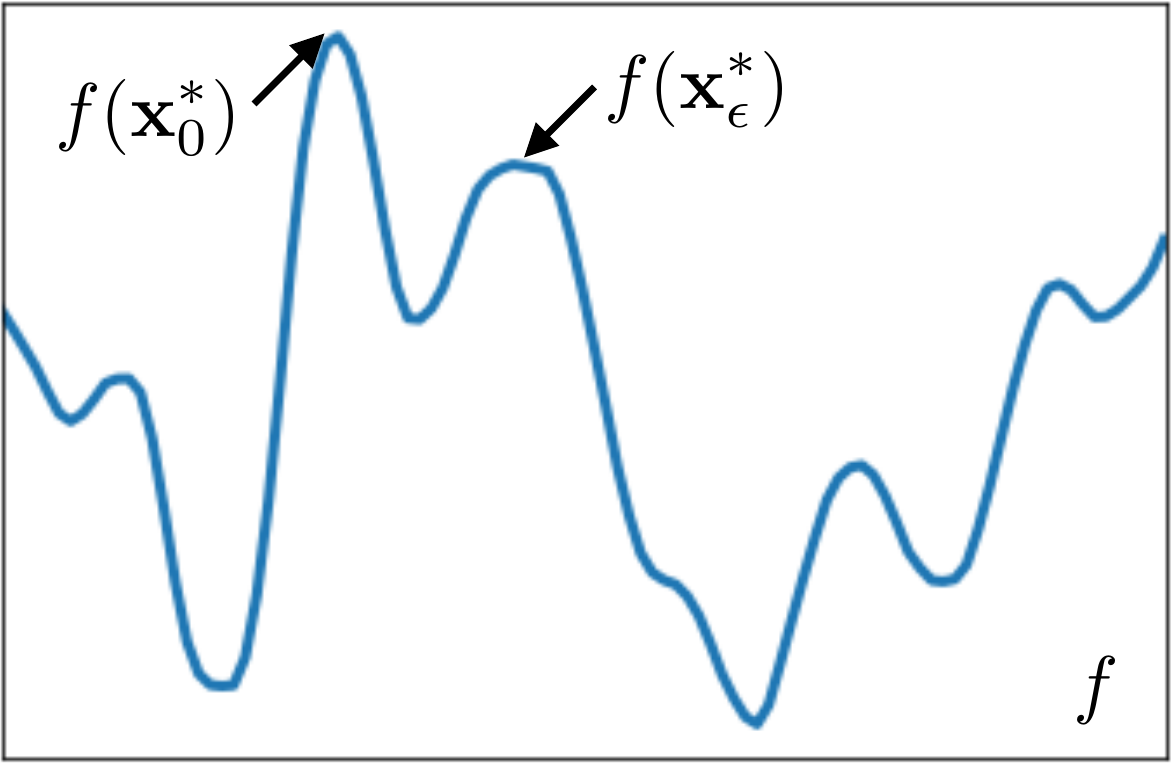}
            \end{subfigure}
            \begin{subfigure}{.32\textwidth}
              \centering
              \includegraphics[scale=0.35]{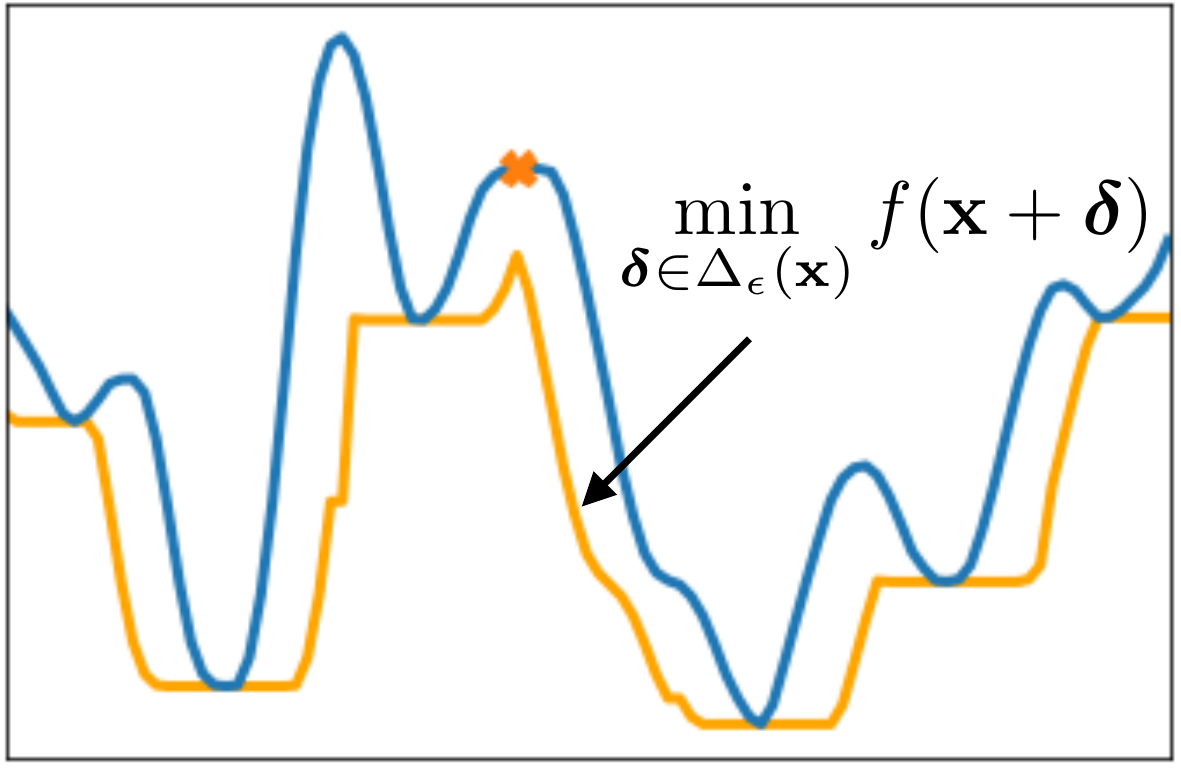}
            \end{subfigure}
            \begin{subfigure}{.32\textwidth}
              \centering
              \includegraphics[scale=0.35]{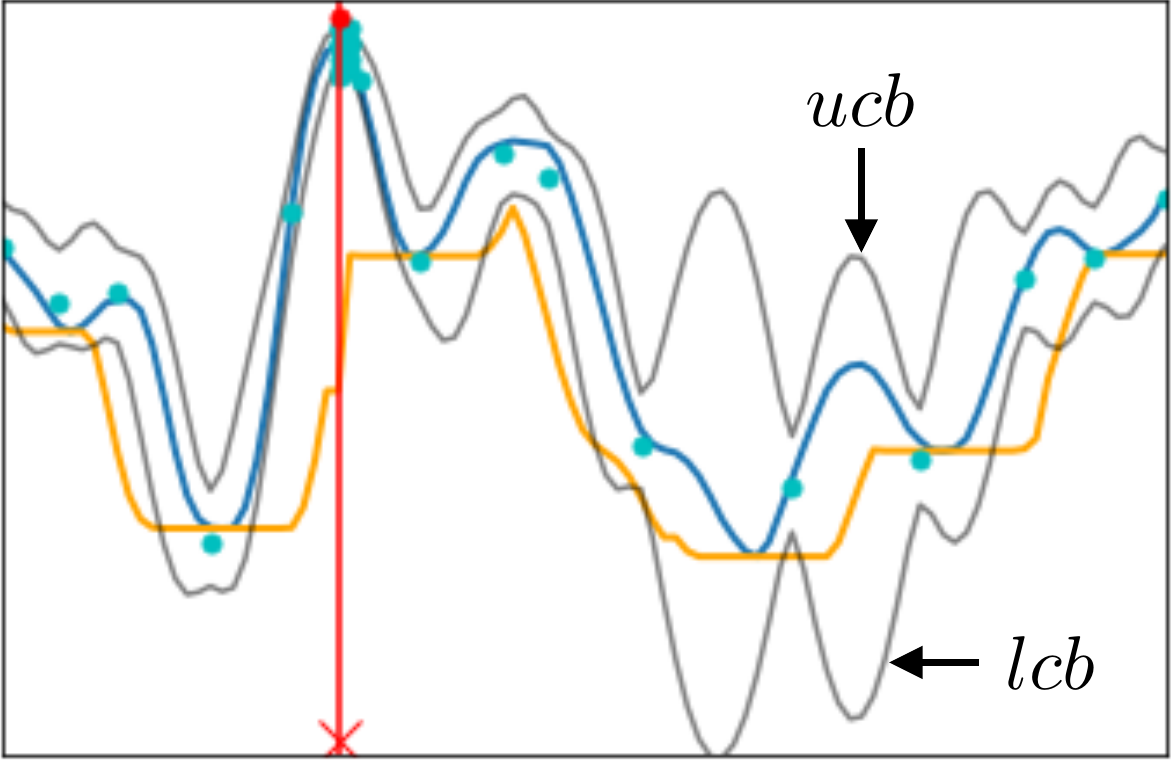}
            \end{subfigure}
            \caption{ (Left) A function $f$ and its maximizer $\x_{0}^*$. (Middle) For $\epsilon= 0.06$ and $d(x,x') = |x-x'|$, the decision that corresponds to the local ``wider'' maximum of $f$ is the \emph{optimal $\epsilon$-stable} decision. (Right) \gpucb~selects a point that nearly maximizes $f$, but is suboptimal in the $\epsilon$-stable sense.}
            \label{fig:problem_formulation}
	\end{figure}

As a running example, we consider the case that $d(\x,\x') = \|\x - \x'\|$ for some norm $\|\cdot\|$ (e.g., $\ell_2$-norm), in which case achieving low $\epsilon$-regret amounts to favoring {\em broad peaks} instead of narrow ones, particularly for higher $\epsilon$; see Figure~\ref{fig:problem_formulation} for an illustration.  In Section \ref{sec:variations}, we discuss how our framework also captures a variety of other max-min optimization settings of interest.

	\textbf{Failure of classical methods.} Various algorithms have been developed for achieving small regret in the standard GP optimization problem. A prominent example is GP-UCB, which chooses 
	\begin{equation}
     \label{eq:gpucb}
	   \x_t \in \argmax_{\x \in D} \ucb_{t-1}(\x),
    \end{equation}
       where $\ucb_{t-1}(\x) := \mu_{t-1}(\x) + \beta_t^{1/2} \sigma_{t-1}(\x)$. This algorithm is guaranteed to achieve sublinear cumulative regret with high probability \cite{srinivas2009gaussian}, for a suitably chosen $\beta_t$. While this is useful when $\x^*_{\epsilon} = \x^*_{0}$,\footnote{In this discussion, we take $d(\x,\x') = \|\x - \x'\|_2$, so that $\epsilon = 0$ recovers the standard non-stable regret \cite{srinivas2009gaussian}.} in general for a given fixed $\epsilon \neq 0$, these two decisions may not coincide, and hence, $\min_{\vdelta \in \Delta_{\epsilon}(\x^*_{0})} f(\x^*_{0} + \vdelta)$ can be significantly smaller than $\min_{\vdelta \in \Delta_{\epsilon}(\x^*_{\epsilon})} f(\x^*_{\epsilon} + \vdelta)$.  
       
       A visual example is given in Figure \ref{fig:problem_formulation} (Right), where the selected point of \gpucb~for $t=20$ is shown. This point nearly maximizes $f$, but it is strictly suboptimal in the $\epsilon$-stable sense. The same limitation applies to other GP optimization strategies (e.g.,~\cite{hennig2012entropy, hernandez2014predictive, bogunovic2016truncated, wang2017max, shekhar2017gaussian, ru2017fast}) whose goal is to identify the global non-robust maximum $\x^*_{0}$. In Section \ref{sec:experiments}, we will see that more advanced baseline strategies also perform poorly when applied to our problem.
%!TEX root = main.tex

\section{Proposed Algorithm and Theory}

Our proposed algorithm, \alg, is described in Algorithm \ref{alg:stable_opt}, and makes use of the following confidence bounds depending on an {\em exploration parameter} $\beta_t$ ({\em cf.}, Lemma \ref{confidence_lemma} below):
\begin{gather}
    \ucb_{t-1}(\x) := \mu_{t-1}(\x) + \beta_t^{1/2} \sigma_{t-1}(\x), \\ \lcb_{t-1}(\x) := \mu_{t-1}(\x) - \beta_t^{1/2} \sigma_{t-1}(\x). \label{eq:ucb_lcb_def}
\end{gather}
The point $\xtil_t$ defined in \eqref{eq:robust_ucb_rule} is the one having the highest ``stable'' upper confidence bound.  However, the queried point is not $\xtil_t$, but instead $\xtil_t + \vdelta_t$, where $\vdelta_t \in \Delta_{\epsilon}(\xtil_t)$ is chosen to minimize the {\em lower} confidence bound.  As a result, the algorithm is based on two distinct principles: (i) optimism in the face of uncertainty when it comes to selecting $\xtil_t$; (ii) pessimism in the face of uncertainty when it comes to anticipating the perturbation of $\xtil_t$.  The first of these is inherent to existing algorithms such as GP-UCB \cite{srinivas2009gaussian}, whereas the second is unique to the adversarially robust GP optimization problem. An example illustration of \alg's execution is given in the supplementary material.

We have left the final reported point $\x^{(T)}$ unspecified in Algorithm \ref{alg:stable_opt}, as there are numerous reasonable choices.  The simplest choice is to simply return $\x^{(T)} = \xtil_T$, but in our theory and experiments, we will focus on $\x^{(T)}$ equaling the point in $\{\xtil_1,\dotsc,\xtil_T\}$ with the highest lower confidence bound.

Finding an exact solution to the optimization of the acquisition function in \eqref{eq:robust_ucb_rule} can be challenging in practice.  When $D$ is continuous, a natural approach is to find an approximate solution using an efficient local search algorithm for robust optimization with a fully known objective function, such as that of \cite{bertsimas2010robust}. % can be used to solve the optimization problem in \ref{eq:robust_ucb_rule} when $D$ is continuous. 

\begin{algorithm}[!t]
    \caption{The \alg ~algorithm} \label{alg:stable_opt}
    \begin{algorithmic}[1]
        \Require Domain $D$, GP prior ($\mu_0$, $\sigma_0$, $k$), parameters $\lbrace \beta_t \rbrace_{t \ge 1}$, stability $\epsilon$, distance function $d(\cdot,\cdot)$
        \For {$t = 1,2,\dotsc, T$}
        \State Set 
        \begin{equation} 
        \xtil_t = \argmax_{\x \in D} \min_{\vdelta \in \Delta_{\epsilon}(\x)} \ucb_{t-1}(\x + \vdelta). \label{eq:robust_ucb_rule} 
        \end{equation}
        \State Set $\vdelta_t = \argmin_{\vdelta \in \Delta_{\epsilon}(\xtil_t)} \;\lcb_{t-1}(\xtil_t + \vdelta)$
        \State Sample $\xtil_t + \vdelta_t$, and observe $y_t = f(\xtil_t + \vdelta_t) + z_t$  
        \State Update $\mu_t$, $\sigma_t$, $\ucb_t$ and $\lcb_t$ according to \eqref{eq:mu_and_sigma_update} and \eqref{eq:ucb_lcb_def}, by including $\lbrace (\xtil_t + \vdelta_t, y_t) \rbrace$ 
        \EndFor
        \State \textbf{end for}
    \end{algorithmic}
\end{algorithm}

\subsection{Upper bound on $\epsilon$-regret}

Our analysis makes use of the {\em maximum information gain} under $t$ noisy measurements: 
\begin{equation}
\gamma_t = \max_{\x_1, \cdots, \x_t} \frac{1}{2} \log \det (\I_t + \sigma^{-2} \vK_t), \label{eq:gamma_def}
\end{equation}
which has been used in numerous theoretical works on GP optimization following \cite{srinivas2009gaussian}.

\alg~depends on the exploration parameter $\beta_t$, which determines the width of the confidence bounds. In our main result, we set $\beta_t$ as in~\cite{chowdhury17kernelized} and make use of the following.

\begin{lemma}
    {\em \cite{chowdhury17kernelized}}
    \label{confidence_lemma}
    Fix $f \in \Fc_k(B)$, and consider the sampling model $y_t = f(\x_t) + z_t$ with $z_t \sim \Normal(0,\sigma^2)$, with independence between times.
    % and (ii) the noise $z_t$ is zero-mean conditioned on the history, and uniformly bounded by some constant $\sigma_0$ for all $t$. Fix $\xi \in (0,1)$ and set $\beta_t = B + \sigma\sqrt{2(\gamma_{t-1} + \log\frac{e}{\delta})}$, 
    %$\beta_t = 2B + 300 \gamma_t \log^3(t \xi^{-1})$, 
    Under the choice $\beta_t = \big(B + \sigma\sqrt{2(\gamma_{t-1} + \log\frac{e}{\xi})}\big)^2$, the following holds with probability at least $1 - \xi$:
    \begin{equation}    
        \lcb_{t-1}(\x) \leq f(\x) \leq \ucb_{t-1}(\x), \quad \forall \x \in D, \forall t \geq 1. \label{eq:conf_bounds}
    \end{equation}

\end{lemma}

The following theorem bounds the performance of \alg~under a suitable choice of the recommended point $\x^{(T)}$.  The proof is given in the supplementary material.

% When $D$ is finite $\gamma_t=\Ord(|D|\log t |D|)$~\cite{srinivas2009gaussian}.

\begin{theorem} \label{thm:upper}
    \emph{(Upper Bound)}
    Fix $\epsilon > 0$, $\eta > 0$, $B > 0$, $T \in \ZZ$, $\xi \in (0,1)$, and a distance function $d(\x,\x')$, and suppose that
    \begin{equation}
        \frac{T}{\beta_T \gamma_T} \geq \frac{C_1}{\eta^2},
    \end{equation}
    where $C_1 = 8 / \log(1 + \sigma^{-2})$.  
    For any $f \in \Fc_k(B)$, \alg~with $\beta_t$ set as in Lemma~\ref{confidence_lemma} achieves $r_{\epsilon}(\x^{(T)}) \le \eta$ after $T$ rounds
%    \begin{equation}
%        \min_{\vdelta \in \Delta_{\epsilon}(\x^{(T)})} f(\x^{(T)} + \vdelta) \geq f(\x^*_{\epsilon}) - \eta,  \label{eq:upper_bound}
%    \end{equation}
    with probability at least $1-\xi$, where 
    \begin{equation}
        \x^{(T)} = \xtil_{t^*}, \qquad t^* = \argmax_{t = 1 ,\dotsc, T} \min_{\vdelta \in \Delta_{\epsilon}(\xtil_t)}  \lcb_{t-1}(\xtil_t + \vdelta). \label{eq:final_point}
    \end{equation}
\end{theorem}

This result holds for general kernels, and for both finite and continuous $D$.  Our analysis bounds function values according to the confidence bounds in Lemma \ref{confidence_lemma} analogously to GP-UCB \cite{srinivas2009gaussian}, but also addresses the non-trivial challenge of characterizing the perturbations $\vdelta_t$.  While we focused on the non-Bayesian RKHS setting, the proof can easily be adapted to handle the {\em Bayesian optimization} (BO) setting in which $f \sim \mathrm{GP}(0,k)$; see Section \ref{sec:variations} for further discussion.

Theorem \ref{thm:upper} can be made more explicit by substituting bounds on $\gamma_T$; in particular, $\gamma_T = O( (\log T)^{p+1} )$ for the SE kernel, and $\gamma_T= O( T^{ \frac{p(p+1)}{ 2\nu + p(p+1) } } \log T )$  for the Mat\'ern-$\nu$ kernel \cite{srinivas2009gaussian}.  The former yields $T = O^*\big( \frac{1}{\eta^2} \big( \log\frac{1}{\eta} \big)^{2p} \big)$ in Theorem~\ref{thm:upper} for constant $B$, $\sigma^2$, and $\epsilon$ (where $O^*(\cdot)$ hides dimension-independent log factors), which we will shortly see nearly matches an algorithm-independent lower bound.

\subsection{Lower bound on $\epsilon$-regret}

Establishing lower bounds under general kernels and input domains is an open problem even in the non-robust setting.  Accordingly, the following theorem focuses on a more specific setting than the upper bound: We let the input domain be $[0,1]^p$ for some dimension $p$, and we focus on the SE and Mat\'ern kernels.  In addition, we only consider the case that $d(\x,\x') = \|\x - \x'\|_2$, though extensions to other norms (e.g., $\ell_{1}$ or $\ell_{\infty}$) follow immediately from the proof.

\begin{theorem} \label{thm:lower}
    \emph{(Lower Bound)}
    Let $D = [0,1]^p$ for some dimension $p$, and set $d(\x,\x') = \|\x - \x'\|_2$.  Fix $\epsilon \in \big(0,\frac{1}{2}\big)$, $\eta \in \big(0,\frac{1}{2}\big)$, $B > 0$, and $T \in \ZZ$.   Suppose there exists an algorithm that, for any $f \in \Fc_k(B)$, reports a point $\x^{(T)}$ achieving $\epsilon$-regret $r_{\epsilon}(\x^{(T)}) \le \eta$ after $T$ rounds with probability at least $1 - \xi$.  Then, provided that $\frac{\eta}{B}$ and $\xi$ are sufficiently small, we have the following:
    \begin{enumerate}[leftmargin=5ex,itemsep=0ex,topsep=0.2ex]
        \item For \emph{$k = \kSE$}, it is necessary that
        % \begin{equation}
        $T = \Omega\big( \frac{\sigma^2}{\eta^2} \big(\log\frac{B}{\eta}\big)^{p/2} \big)$. % \label{eq:inst_se}
        % \end{equation}
        \item For \emph{$k = \kMat$}, it is necessary that
        % \begin{equation}
        $T = \Omega\big( \frac{\sigma^2}{\eta^2} \big(\frac{B}{\eta}\big)^{p/\nu} \big).$ % \label{eq:inst_Mat}
        % \end{equation}
    \end{enumerate}
    Here we assume that the stability parameter $\epsilon$, dimension $p$, target probability $\xi$, and kernel parameters $l,\nu$ are fixed (i.e., not varying as a function of the parameters $T$, $\eta$, $\sigma$ and $B$).
\end{theorem}

The proof is based on constructing a finite subset of ``difficult'' functions in $\Fc_k(B)$ and applying lower bounding techniques from the multi-armed bandit literature, also making use of several auxiliary results from the non-robust setting \cite{scarlett2017lower}.  More specifically, the functions in the restricted class consist of narrow negative ``valleys'' that the adversary can perturb the reported point into, but that are hard to identify until a large number of samples have been taken.

For constant $\sigma^2$ and $B$, the condition for the SE kernel simplifies to $T = \Omega\big( \frac{1}{\eta^2} \big(\log\frac{1}{\eta}\big)^{p/2} \big)$, thus nearly matching the upper bound $T = O^*\big( \frac{1}{\eta^2} \big( \log\frac{1}{\eta} \big)^{2p} \big)$ of \alg~established above. In the case of the Mat\'ern kernel, more significant gaps remain between the upper and lower bounds; however, similar gaps remain even in the standard (non-robust) setting \cite{scarlett2017lower}.

\section{Variations of \alg} \label{sec:variations}

While the above problem formulation seeks robustness within an $\epsilon$-ball corresponding to the distance function $d(\cdot,\cdot)$, our algorithm and theory apply to a variety of seemingly distinct settings.  We outline a few such settings here; in the supplementary material, we give details of their derivations.

\textbf{Robust Bayesian optimization.} Algorithm \ref{alg:stable_opt} and Theorem \ref{thm:upper} extend readily to the Bayesian setting in which $f \sim \mathrm{GP}(0, k(\x,\x'))$.  In particular, since the proof of Theorem \ref{thm:upper} is based on confidence bounds, the only change required is selecting $\beta_t$ to be that used for the Bayesian setting in \cite{srinivas2009gaussian}.  As a result, our framework also captures the novel problem of {\em adversarially robust Bayesian optimization}.  All of the variations outlined below similarly apply to both the Bayesian and non-Bayesian settings.

\textbf{Robustness to unknown parameters.}  Consider the scenario where an unknown function $f:D \times \Theta \rightarrow \mathbb{R}$ has a bounded RKHS norm under some composite kernel $k((\x,\vtheta), (\x',\vtheta'))$. Some special cases include $k((\x,\vtheta), (\x',\vtheta')) = k(\x,\x') + k(\vtheta, \vtheta')$ and $k((\x,\vtheta), (\x',\vtheta')) = k(\x,\x')k(\vtheta, \vtheta')$ \cite{krause2011contextual}.  The posterior mean $\mu_{t}(\x, \vtheta)$ and variance $\sigma^2_{t}(\x, \vtheta)$ conditioned on the previous observations $(\x_1, \vtheta_1, y_1), ..., (\x_{t-1}, \vtheta_{t-1}, y_{t-1})$ are computed analogously to \eqref{eq:mu_and_sigma_update} \cite{krause2011contextual}. 

A robust optimization formulation in this setting is to seek $\x$ that solves
\begin{equation}
    \label{eq:robust_opt}
    \max_{\x \in D} \min_{\vtheta \in \Theta} f(\x, \vtheta).
\end{equation}
That is, we seek to find a configuration $\x$ that is robust against any possible parameter vector $\vtheta \in \Theta$.  

Potential applications of this setup include the following:
\begin{itemize}[leftmargin=5ex,itemsep=0ex,topsep=0.25ex]
    \item In areas such a robotics, we may have numerous parameters to tune (given by $\x$ and $\vtheta$ collectively), but when it comes to implementation, some of them (i.e., only $\vtheta$) become out of our control.  Hence, we need to be robust against whatever values they may take.
    \item We wish to tune hyperparameters in order to make an algorithm work simultaneously for a number of distinct data types that bear some similarities/correlations.  The data types are represented by $\vtheta$, and we can choose the data type to our liking during the optimization stage.
\end{itemize} 
\alg~can be used to solve \eqref{eq:robust_opt}; we maintain $\vtheta_t$ instead of $\vdelta_t$, and modify the main steps to
\begin{gather}
  \x_t \in \argmax_{\x \in D} \min_{\vtheta \in \Theta} \ucb_{t-1}(\x, \vtheta), \\ % \;\; \text{ and } \;\;
  \vtheta_t \in \argmin_{\vtheta \in \Theta} \lcb_{t-1} (\x_t, \vtheta). \label{eq:variation1}
\end{gather} 
At each time step, \alg~receives a noisy observation $y_t = f(\x_t, \vtheta_t) + z_t$, which is used with $(\x_t, \vtheta_t)$ for computing the posterior.  As explained in the supplementary material, the guarantee $r_{\epsilon}(\x^{(T)}) \le \eta$ in Theorem \ref{thm:upper} is replaced by $\min_{\vtheta \in \Theta} f(\x^{(T)}, \vtheta) \geq \max_{\x \in D} \min_{\vtheta \in \Theta} f(\x, \vtheta) - \eta$.

\textbf{Robust estimation.} Continuing with the consideration of a composite kernel on $(\x,\vtheta)$, we consider the following practical problem variant proposed in~\cite{bertsimas2010robust}. Let $\bar{\vtheta} \in \Theta$ be an estimate of the true problem coefficient $\vtheta^* \in \Theta$. Since, $\bar{\vtheta}$ is an estimate, the true coefficient satisfies $\vtheta^* = \bar{\vtheta} + \vdelta_{\vtheta}$, where $\vdelta_{\vtheta}$ represents uncertainty in $\bar{\vtheta}$.   Often, practitioners solve $\max_{\x \in D} f(\x, \bar{\vtheta})$ and ignore the uncertainty. As a more sophisticated approach, we let $\Delta_{\epsilon}(\bar{\vtheta}) = \big\{ \vtheta - \bar{\vtheta}:\, \vtheta \in \Theta \; \text{and}\; d(\bar{\vtheta},\vtheta) \le \epsilon \big\}$, and consider the following robust problem formulation:

\begin{equation}
    \max_{\x \in D} \min_{\vdelta_{\vtheta} \in \Delta_{\epsilon}(\bar{\vtheta})} f(\x, \bar{\vtheta} + \vdelta_{\vtheta}).
\end{equation} 
For the given estimate $\bar{\vtheta}$, the main steps of \alg~in this setting are
\begin{gather}
  \x_t \in \argmax_{\x \in D} \min_{\vdelta_{\vtheta} \in \Delta_{\epsilon}(\bar{\vtheta})} \ucb_{t-1}(\x, \bar{\vtheta}+\vdelta_{\vtheta}), \\ %  \;\; \text{ and } \;\;  
  \vdelta_{\vtheta,t} \in \argmin_{\vdelta_{\vtheta} \in \Delta_{\epsilon}(\bar{\vtheta})} \lcb_{t-1} (\x_t, \bar{\vtheta}+\vdelta_{\vtheta}), \label{eq:variation2}
\end{gather}
and the noisy observations take the form $y_t = f(\x_t, \bar{\vtheta}+\vdelta_{\vtheta,t}) + z_t$.  The guarantee $r_{\epsilon}(\x^{(T)}) \le \eta$ in Theorem \ref{thm:upper} is replaced by $\min_{\vdelta_{\vtheta} \in \Delta_{\epsilon}(\bar{\vtheta})} f(\x^{(T)}, \bar{\vtheta}+\vdelta_{\vtheta}) \geq \max_{\x \in D} \min_{\vdelta_{\vtheta} \in \Delta_{\epsilon}(\bar{\vtheta})} f(\x, \bar{\vtheta}+\vdelta_{\vtheta}) - \eta$.

\textbf{Robust group identification.}  In some applications, it is natural to partition $D$ into disjoint subsets, and search for the subset with the highest worst-case function value (see Section \ref{sec:experiments} for a movie recommendation example).  Letting $\mathcal{G} = \lbrace G_1, \dots, G_k \rbrace$ denote the groups that partition the input space, the robust optimization problem is given by 
\begin{equation}
    \max_{G \in \mathcal{G}} \min_{\x \in G} f(\x),
\end{equation}
and the algorithm reports a group $G^{(T)}$.  The main steps of \alg~are given by
\begin{gather}
G_t \in \argmax_{G \in \mathcal{G}} \min_{\x \in G} \ucb_{t-1}(\x), \\ %  \;\; \text{ and } \;\;  
\x_{t} \in \argmin_{\x \in G_t} \lcb_{t-1} (\x), \label{eq:variation3}
\end{gather}  
and the observations are of the form $y_t = f(\x_t) + z_t$ as usual.  The guarantee $r_{\epsilon}(\x^{(T)}) \le \eta$ in Theorem \ref{thm:upper} is replaced by $\min_{\x \in G^{(T)}} f(\x) \geq \max_{G \in \mathcal{G}} \min_{\x \in G} f(\x) - \eta$.

The preceding variations of \alg, as well as their resulting variations of Theorem \ref{thm:upper},
 follow by substituting certain (unconventional) choices of $d(\cdot,\cdot)$ and $\epsilon$ into Algorithm \ref{alg:stable_opt} and Theorem \ref{thm:upper}, with $(\x,\vtheta)$ in place of $\x$ where appropriate. The details are given in the supplementary material. 
%!TEX root = main.tex

\section{Experiments} \label{sec:experiments}

In this section, we experimentally validate the performance of \alg~by comparing against several baselines.  Each algorithm that we consider may distinguish between the {\em sampled point} (i.e., the point that produces the noisy observation $y_t$) and the {\em reported point} (i.e., the point believed to be near-optimal in terms of $\epsilon$-stability).  For \alg, as described in Algorithm \ref{alg:stable_opt}, the sampled point is $\xtil_t + \vdelta_t$, and the reported point after time $t$ is the one in $\lbrace \xtil_1, \dots, \xtil_t \rbrace$ with the highest value of $\min_{\vdelta \in \Delta_{\epsilon}(\xtil_t)} \lcb_t(\xtil_t + \vdelta)$.\footnote{This is slightly different from Theorem \ref{thm:upper}, which uses the confidence bound $\lcb_{\tau-1}$ for $\x_{\tau}$ instead of adopting the common bound $\lcb_t$.  We found the latter to be more suitable when the kernel hyperparameters are updated online, whereas Theorem \ref{thm:upper} assumes a known kernel.  Theorem \ref{thm:upper} can be adapted to use $\lcb_t$ alone by intersecting the confidence bounds at each time instant so that they are monotonically shrinking \cite{gotovos2013active}. }  In addition, we consider the following baselines:
\begin{itemize}[leftmargin=5ex]
  \item \gpucb~(see \eqref{eq:gpucb}). We consider \gpucb~ to be a good representative of the wide range of existing methods that search for the non-robust global maximum. 
  \item \mmgpucb. We consider a natural generalization of GP-UCB in which, at each time step, the sampled and reported point are both given by 
  \begin{equation}
    \x_t = \argmax_{\x \in D} \min_{\vdelta \in \Delta_{\epsilon}(x)} \ucb_{t-1}(\x + \vdelta).
  \end{equation}
  \item \srandom. The sampling point $\x_t$ at every time step is chosen uniformly at random, while the reported point at time $t$ is chosen to be the point among the sampled points $\lbrace \x_1, \dots, \x_t \rbrace$ according to the same rule as the one used for \alg.
  \item \sgpucb. The sampled point is given by the \gpucb~rule, while the reported point is again chosen in the same way as in \alg.
\end{itemize}  
As observed in existing works (e.g., \cite{srinivas2009gaussian,bogunovic2016truncated}), the theoretical choice of $\beta_t$ is overly conservative.  We therefore adopt a constant value of $\beta_t^{1/2} = 2.0$ in each of the above methods, which we found to provide a suitable exploration/exploitation trade-off for each of the above algorithms.

Given a reported point $\x^{(t)}$ at time $t$, the performance metric is the $\epsilon$-regret $r_{\epsilon}(\x^{(t)})$ given in \eqref{eq:eps_regret}.
Two observations are in order: (i) All the baselines are heuristic approaches for our problem, and they do not have guarantees in terms of near-optimal stability;
(ii) We do not compare against other standard GP optimization methods, as their performance is comparable to that of \gpucb; in particular, they suffer from exactly the same pitfalls described at the end of Section \ref{sec:setup}.

\begin{figure}
            \begin{subfigure}{.33\textwidth}
              \centering
              \includegraphics[scale=0.35]{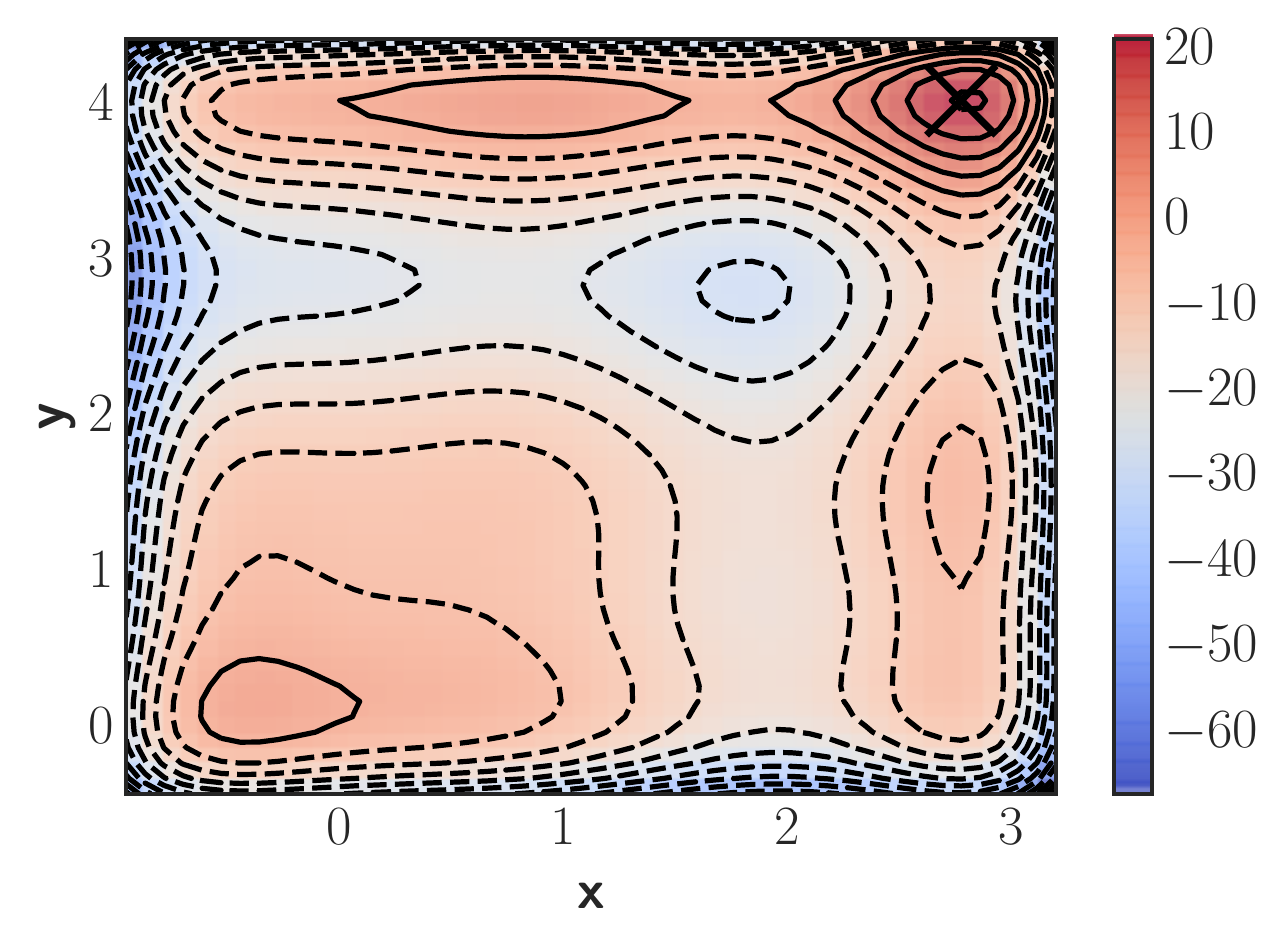}
              \caption{$\fpoly(x,y)$}
              \label{fig:bertsimas_f}
            \end{subfigure}
            \begin{subfigure}{.33\textwidth}
              \centering
              \includegraphics[scale=0.35]{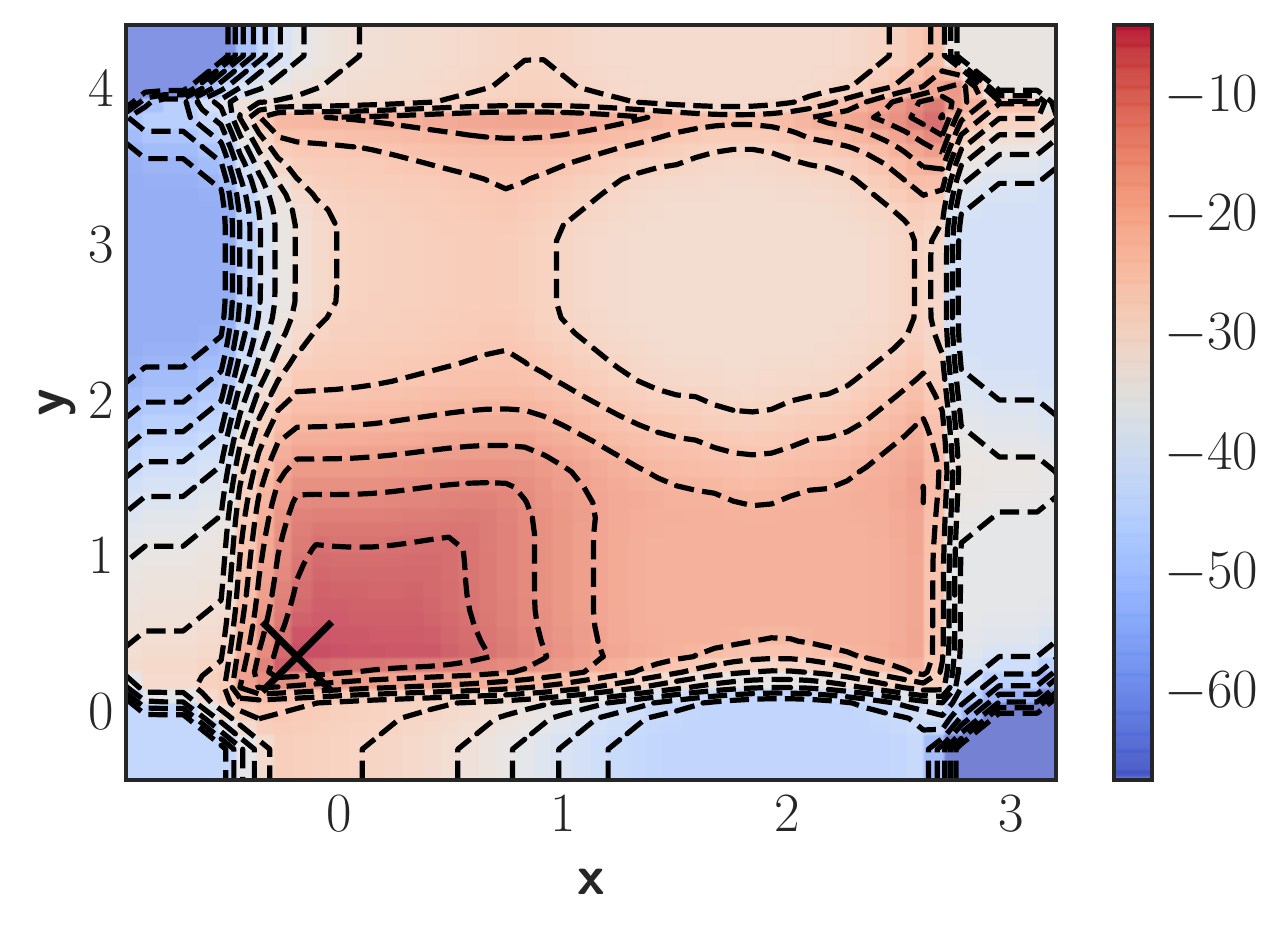}
              \caption{$\gpoly(x,y)$}
              \label{fig:bertsimas_rob_f}
            \end{subfigure}
            \begin{subfigure}{.33\textwidth}
              \centering
              \includegraphics[scale=0.35]{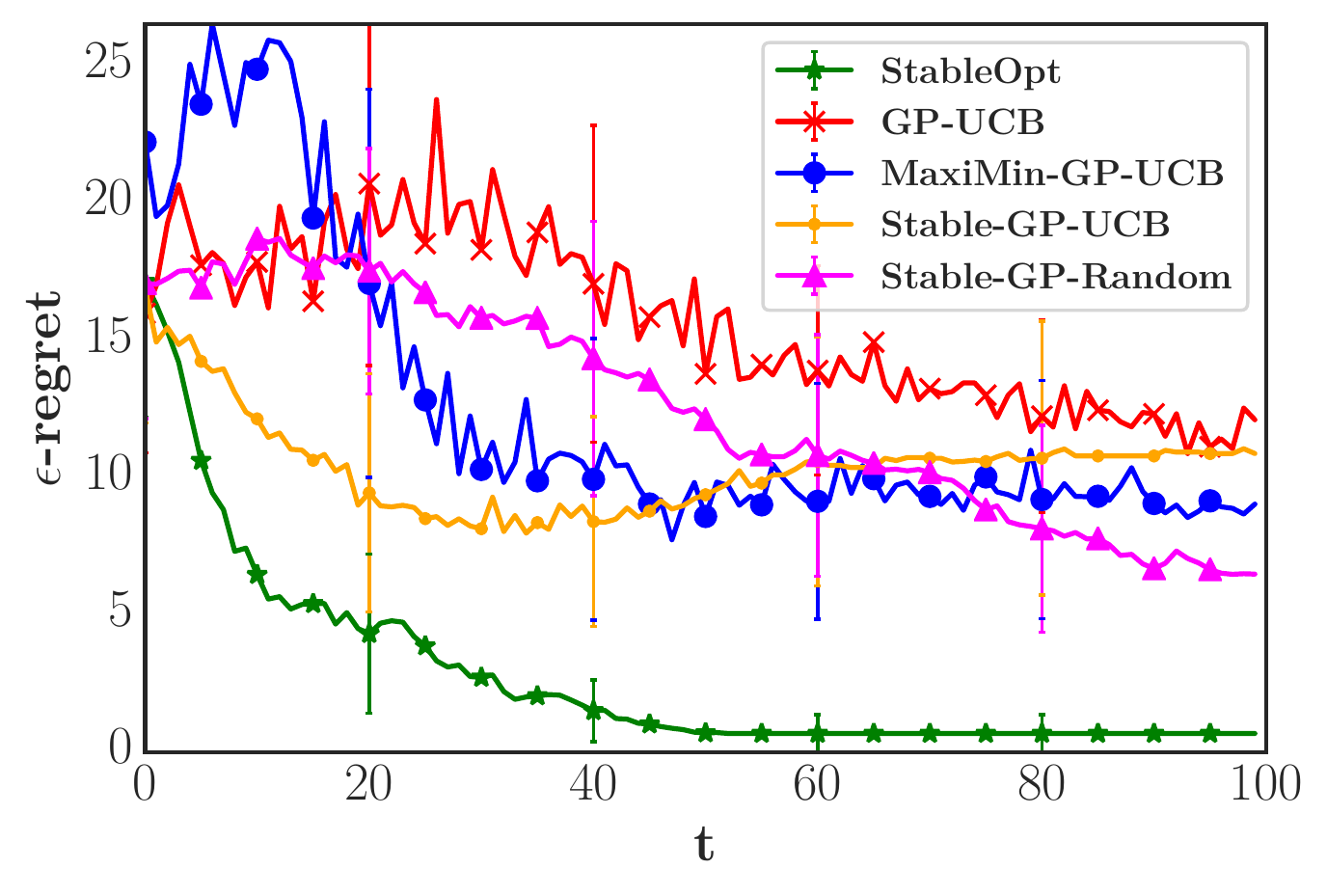}
              \caption{$\epsilon$-regret}
              \label{fig:bertsimas_regret}
            \end{subfigure}
            \caption{(Left) Synthetic function from \cite{bertsimas2010robust}. (Middle) Counterpart with worst-case perturbations. (Right) The performance.  In this example, \alg~significantly outperforms the baselines.}
            \label{fig:synthetic}
            \vspace*{-1ex}  
\end{figure}

\textbf{Synthetic function.}
We consider the synthetic function from~\cite{bertsimas2010robust} (see Figure~\ref{fig:bertsimas_f}), given by  
\begin{align}
  \fpoly(x,y) &= -2x^6 + 12.2 x^5 - 21.2x^4 - 6.2x + 6.4x^3 + 4.7x^2 - y^6 + 11y^5 \nonumber\\ 
         &- 43.3y^4 + 10y + 74.8y^3 - 56.9y^2 + 4.1xy + 0.1y^2x^2 - 0.4y^2x - 0.4x^2y.
\end{align}
The decision space is a uniformly spaced grid of points in $((-0.95, 3.2),(-0.45, 4.4))$ of size $10^4$. There exist multiple local maxima, and the global maximum is at $(x_{f}^*, y_{f}^*) = (2.82, 4.0)$, with $\fpoly(x_{f}^*, y_{f}^*)= 20.82$. Similarly as in~\cite{bertsimas2010robust}, given stability parameters $\vdelta = (\delta_x, \delta_y)$, where $\| \vdelta \|_2 \leq 0.5$, the robust optimization problem is 
\begin{equation}
    \max_{(x,y) \in D} \gpoly(x,y),
\end{equation}
    where
\begin{equation}
    \gpoly(x,y):= \min_{(\delta_x, \delta_y) \in \Delta_{0.5}(x,y)} \fpoly(x - \delta_x, y - \delta_y).
\end{equation}
A plot of $\gpoly$ is shown in Figure~\ref{fig:bertsimas_rob_f}. The global maximum is attained at 
$(x_{g}^*,y_{g}^*)=(-0.195,0.284)$ and $\gpoly(x_{g}^*,y_{g}^*)=-4.33$, and the inputs maximizing $f$ yield $\gpoly(x_{f}^*,y_{f}^*)=-22.34$.

We initialize the above algorithms by selecting $10$ uniformly random inputs $(x,y)$, keeping those points the same for each algorithm. The kernel adopted is a squared exponential ARD kernel. We randomly sample $500$ points whose function value is above $-15.0$ to learn the GP hyperparameters via maximum likelihood, and then run the algorithms with these hyperparameters. The observation noise standard deviation is set to $0.1$, and the number of sampling rounds is $T=100$. We repeat the experiment $100$ times and show the average performance in Figure~\ref{fig:bertsimas_regret}. We observe that \alg~significantly outperforms the baselines in this experiment. In the later rounds, the baselines report points that are close to the global optimizer, which is suboptimal with respect to the $\epsilon$-regret.

\textbf{Lake data.} In the supplementary material, we provide an analogous experiment to that above using chlorophyll concentration data from Lake Z\"urich, with \alg~again performing best.

\textbf{Robust robot pushing.} 
We consider the deterministic version of the robot pushing objective from~\cite{wang2017max}, with publicly available code.\footnote{https://github.com/zi-w/Max-value-Entropy-Search} The goal is to find a good pre-image for pushing an object to a target location. The 3-dimensional function takes as input the robot location $(r_x, r_y)$ and pushing duration $r_t$, and outputs $f(r_x, r_y, r_t) = 5 - d_{\mathrm{end}}$, where $d_{\mathrm{end}}$ is the distance from the pushed object to the target location. The domain $D$ is continuous: $r_x, r_y \in [-5,5]$ and $r_t \in [1,30]$.

We consider a twist on this problem in which there is uncertainty regarding the precise target location, so one seeks a set of input parameters that is robust against a number of different potential locations.  In the simplest case, the number of such locations is finite, meaning we can model this problem as $\vr \in \argmax_{\vr \in D} \min_{i \in [m]} f_i(\vr)$, where each $f_i$ corresponds to a different target location, and $[m] = \{1,\dotsc,m\}$.  This is a special case of \eqref{eq:robust_opt} with a finite set $\Theta$ of cardinality $m$.

\begin{figure}
            \centering
            \begin{subfigure}{.33\textwidth}
              \centering
              \includegraphics[scale=0.35]{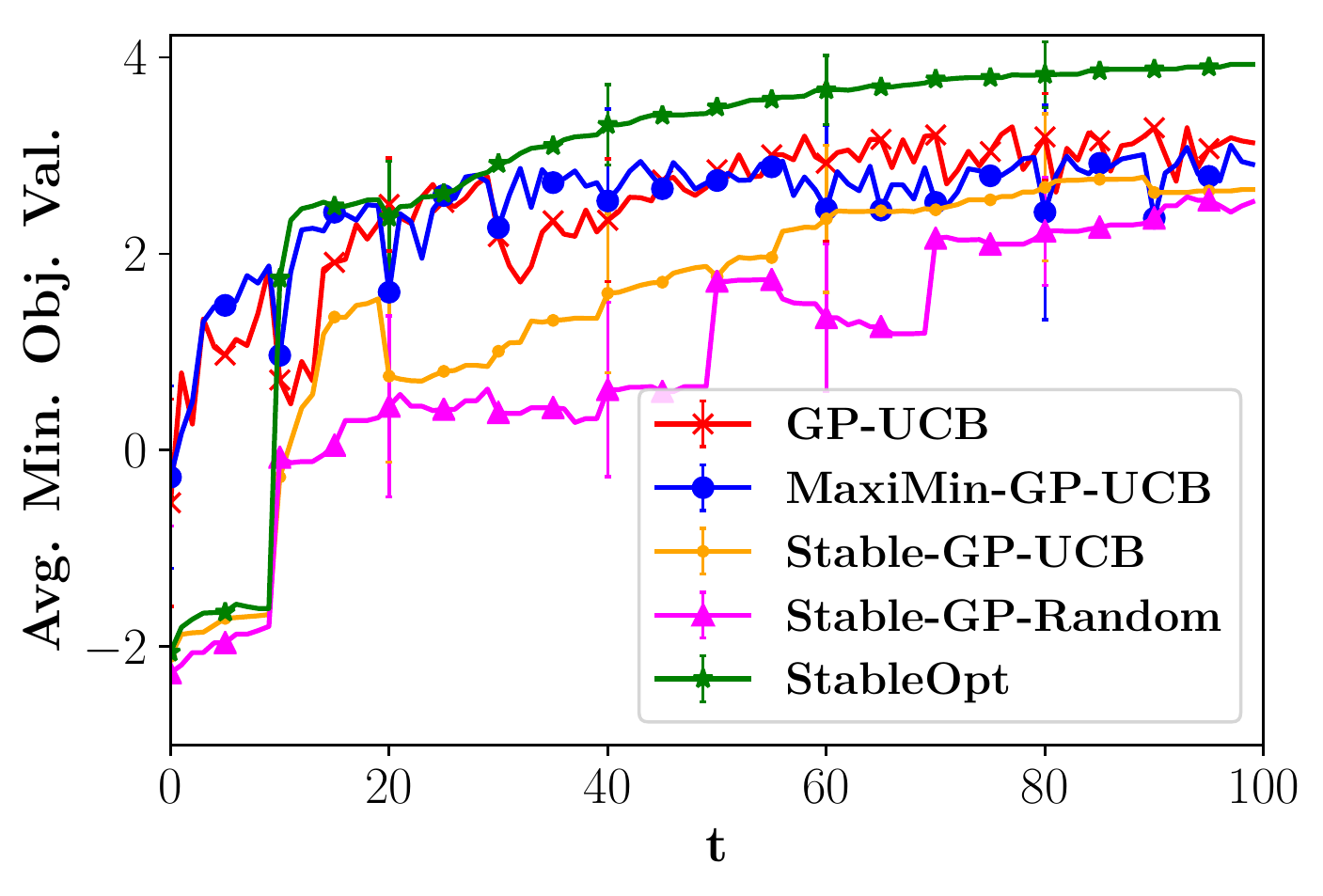}
            \end{subfigure}
            \hspace{2cm}
            \begin{subfigure}{.33\textwidth}
              \centering
              \includegraphics[scale=0.35]{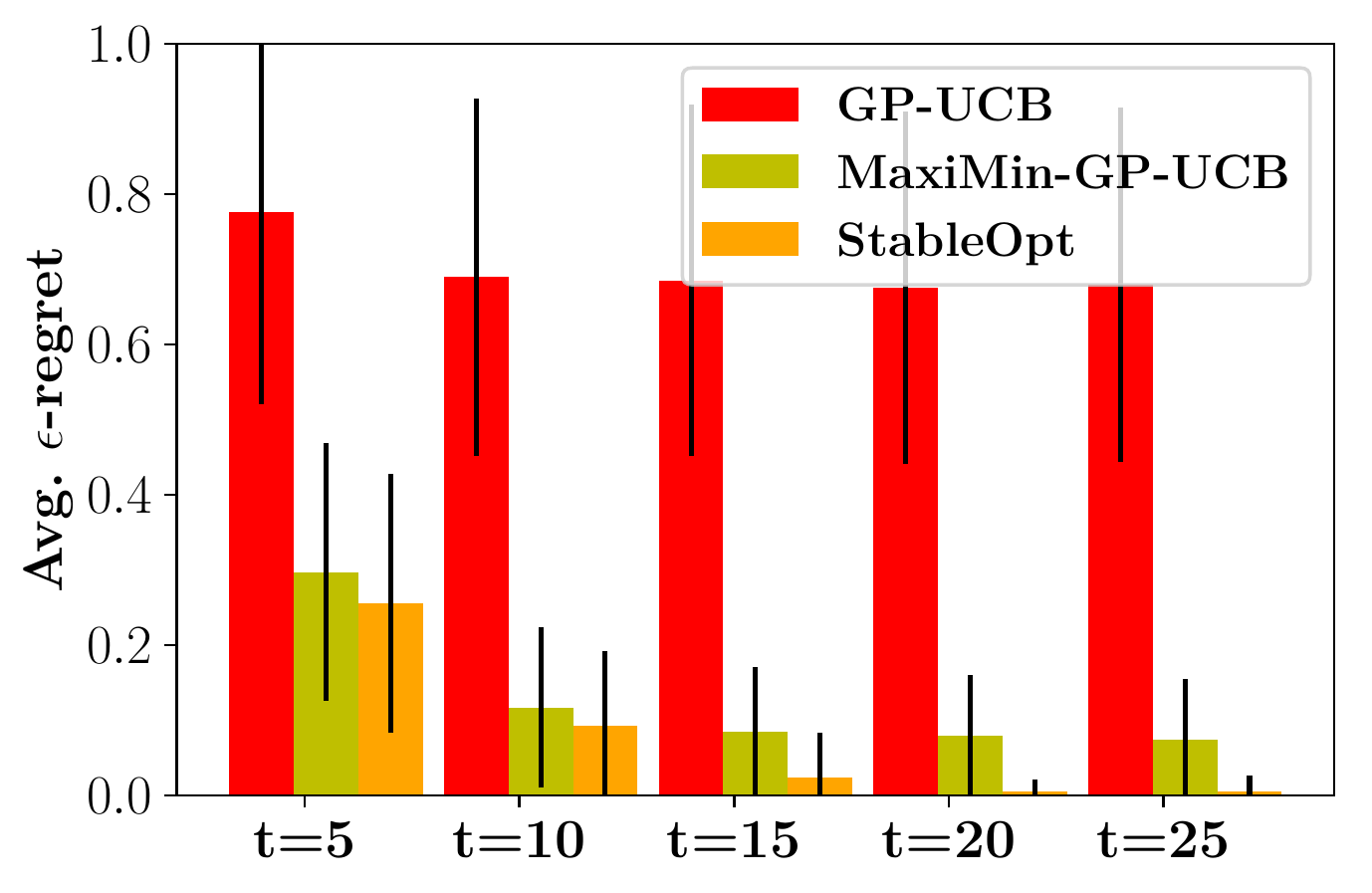}
            \end{subfigure}
            \caption{Robust robot pushing experiment (Left) and MovieLens-100K experiment (Right)}
            \label{fig:rw_applications}
            \vspace*{-1ex}  
\end{figure}

In our experiment, we use $m=2$. Hence, our goal is to find an input configuration $\vr$ that is robust against two different target locations. The first target is uniform over the domain, and the second is uniform over the $\ell_1$-ball centered at the first target location with radius $r=2.0$. We initialize each algorithm with one random sample from each $f_i$. We run each method for $T = 100$ rounds, and for a reported point $\vr_t$ at time $t$, we compare the methods in terms of the robust objective $\min_{i \in [m]} f_{i}(\vr_t)$. We perform a fully Bayesian treatment of the hyperparameters, sampling every $10$ rounds as in~\cite{hernandez2014predictive,wang2017max}. We average over $30$ random pairs of $\lbrace f_1, f_2 \rbrace$ and report the results in Figure~\ref{fig:rw_applications}.   \alg~noticeably outperforms its competitors except in some of the very early rounds.  We note that the apparent discontinuities in certain curves are a result of the hyperparmeter re-estimation.

\textbf{Group movie recommendation.}
Our goal in this task is to recommend a group of movies to a user such that {\em every} movie in the group is to their liking. We use the MovieLens-100K dataset, which consists of 1682 movies and 943 users. The data takes the form of an incomplete matrix $\R$ of ratings, where $R_{i,j}$ is the rating of movie $i$ given by the user $j$. To impute the missing rating values, we apply non-negative matrix factorization with $p=15$ latent factors. This produces a feature vector for each movie $\m_i \in \mathbb{R}^{p}$ and user $\vu_j \in \mathbb{R}^{p}$. We use $10\%$ of the user data for training, in which we fit a Gaussian distribution $P(\vu) = \mathcal{N}(\vu| \boldsymbol{\mu}, \boldsymbol{\Sigma})$. For a given user $\vu_j$ in the test set, $P(\vu)$ is considered to be a prior, and the objective is given by $f_{j}(\m_i) = \m_i^T\vu_{j}$, corresponding to a GP with a linear kernel.

We cluster the movie feature vectors into $k=80$ groups, i.e., $\mathcal{G} = \lbrace G_1, \dots, G_k \rbrace$, via the $k$-means algorithm. Similarly to \eqref{eq:variation3}, the robust optimization problem for a given user $j$ is
\begin{equation}
  \label{eq:group_obj}
  \max_{G \in \mathcal{G}} g_j(G),
\end{equation}
where $g_j(G) = \min_{\m_i \in G} f_{j}(\m_i)$.
That is, for the user with feature vector $\vu_j$, our goal is to find the group of movies to recommend such that the entire collection of movies is robust with respect to the user's preferences.

In this experiment, we compare \alg~against \gpucb~and \mmgpucb. We report the $\epsilon$-regret given by $g_j(G^*) - g_j(G^{(t)})$ %$\min_{\m_i \in G^*} f_{j}(\m_i) - \min_{\m_i \in G_t} f_{j}(\m_i)$, 
where $G^*$ is the maximizer of \eqref{eq:group_obj}, and $G^{(t)}$ is the reported group after time $t$. Since we are reporting groups rather than points, the baselines require slight modifications: At time $t$, \gpucb~selects the movie $\m_t$ (i.e., asks for the user's rating of it) and reports the group $G^{(t)}$ to which $\m_t$ belongs. \mmgpucb~reports $G^{(t)} \in \argmax_{G \in \mathcal{G}} \min_{\m \in G} \ucb_{t-1}(\m)$ and then selects $\m_t \in \argmin_{\m \in G^{(t)}} \ucb_{t-1}(\m)$. Finally, \alg~reports a group in the same way as \mmgpucb, but selects $\m_t$ analogously to \eqref{eq:variation3}. In Figure~\ref{fig:rw_applications}, we show the average $\epsilon$-regret, where the average is taken over $500$ different test users. In this experiment, the average $\epsilon$-regret decreases rapidly after only a small number of rounds. Among the three methods, \alg~is the only one that finds the optimal movie group.    

%!TEX root = main.tex
% \vspace*{-1ex}
\section{Conclusion}
% \vspace*{-1ex}

We have introduced and studied a variant of GP optimization in which one requires stability/robustness to an adversarial perturbation.  We demonstrated the failures of existing algorithms, and provided a new algorithm \alg~that overcomes these limitations, with rigorous guarantees.  We showed that our framework naturally applies to several interesting max-min optimization formulations, and we demonstrated significant improvements over some natural baselines in the experimental examples.

An interesting direction for future work is to study the $\epsilon$-stable optimization formulation in the context of hyperparameter tuning (e.g., for deep neural networks).  One might expect that wide function maxima in hyperparameter space provide better generalization than narrow maxima, but establishing this requires further investigation.  Similar considerations are an ongoing topic of debate in understanding the {\em parameter space} rather than the hyperparmeter space, e.g., see \cite{dinh2017sharp}.

% {\bf Acknowledgment.} This work was supported by the European Research Council
% (ERC) under the European Union’s Horizon 2020 research and innovation programme (grant agreement
% 725594 – time-data), and by an NUS startup grant.

{\bf Acknowledgment.} This work was partially supported by the Swiss National Science Foundation (SNSF) under grant number 407540\_167319, by the European Research Council (ERC) under the European Union's Horizon 2020 research and innovation programme (grant agreement no725594 - time-data), by DARPA DSO’s Lagrange program under grant FA86501827838, and by an NUS startup grant.

% \newpage
\bibliographystyle{plain}
\bibliography{IB_refs}

\newpage
%!TEX root = main.tex
\appendix

{\centering
    {\huge \bf Supplementary Material}
    
    {\Large \bf Adversarially Robust Optimization with Gaussian Processes \\{\small Ilija Bogunovic, Jonathan Scarlett, Stefanie Jegelka and Volkan Cevher (NIPS 2018)} \par }  
}
 
\section{Illustration of \alg's~Execution}
    The following figure gives an example of the selection procedure of \alg~at two different time steps:
    \begin{figure}[H]
            \begin{subfigure}{.5\textwidth}
              \centering
              \includegraphics[scale=0.42]{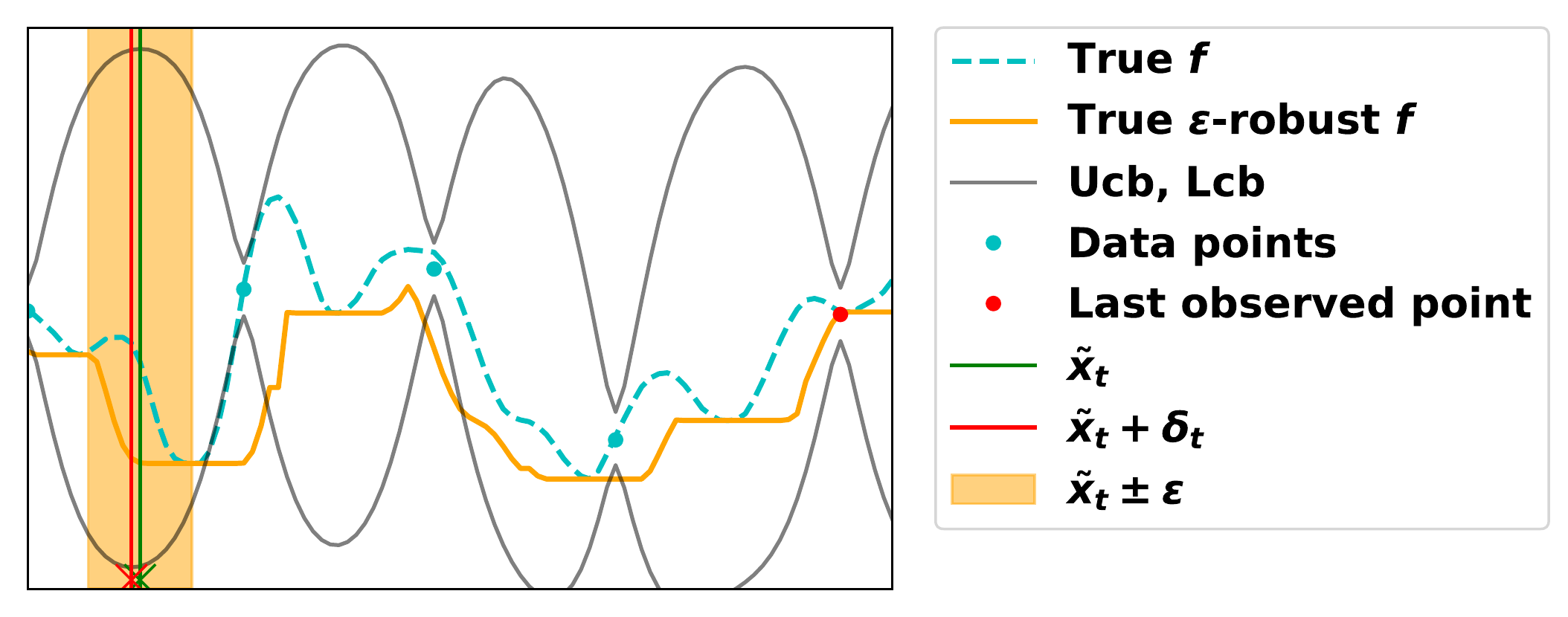}
              \caption{$t=5$}
            \end{subfigure}
            \hspace{1cm}
            \begin{subfigure}{.5\textwidth}
              \centering
              \includegraphics[scale=0.42]{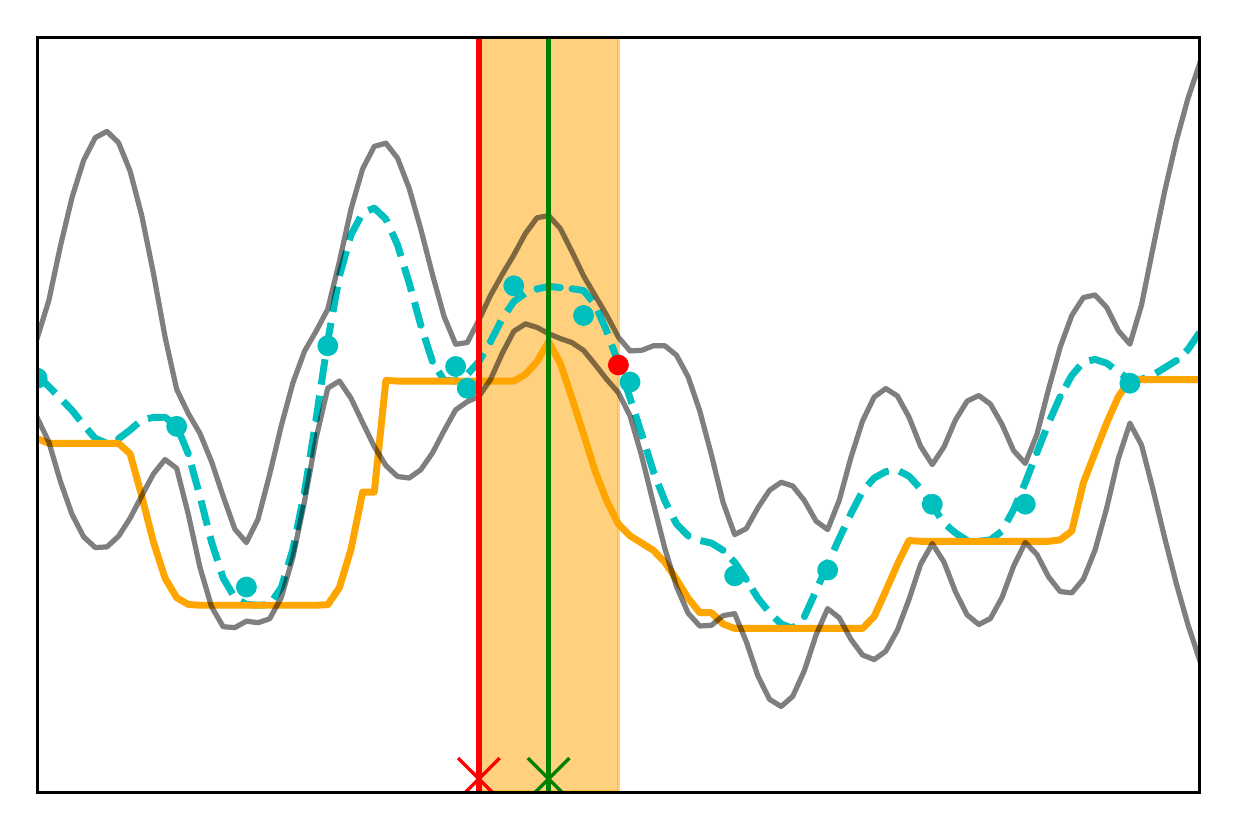}
              \caption{$t=15$}
            \end{subfigure}
            \caption{An execution of \alg~on the running example. We observe that after $t=15$ steps, $\xtil_t$ obtained in Eq.~\ref{eq:robust_ucb_rule} corresponds to $\x_{\epsilon}^*$.}
    \end{figure}
    
    The intermediate time steps are illustrated as follows:
    \begin{figure}[H]
            \begin{subfigure}{.3\textwidth}
              \centering
              \includegraphics[scale=0.35]{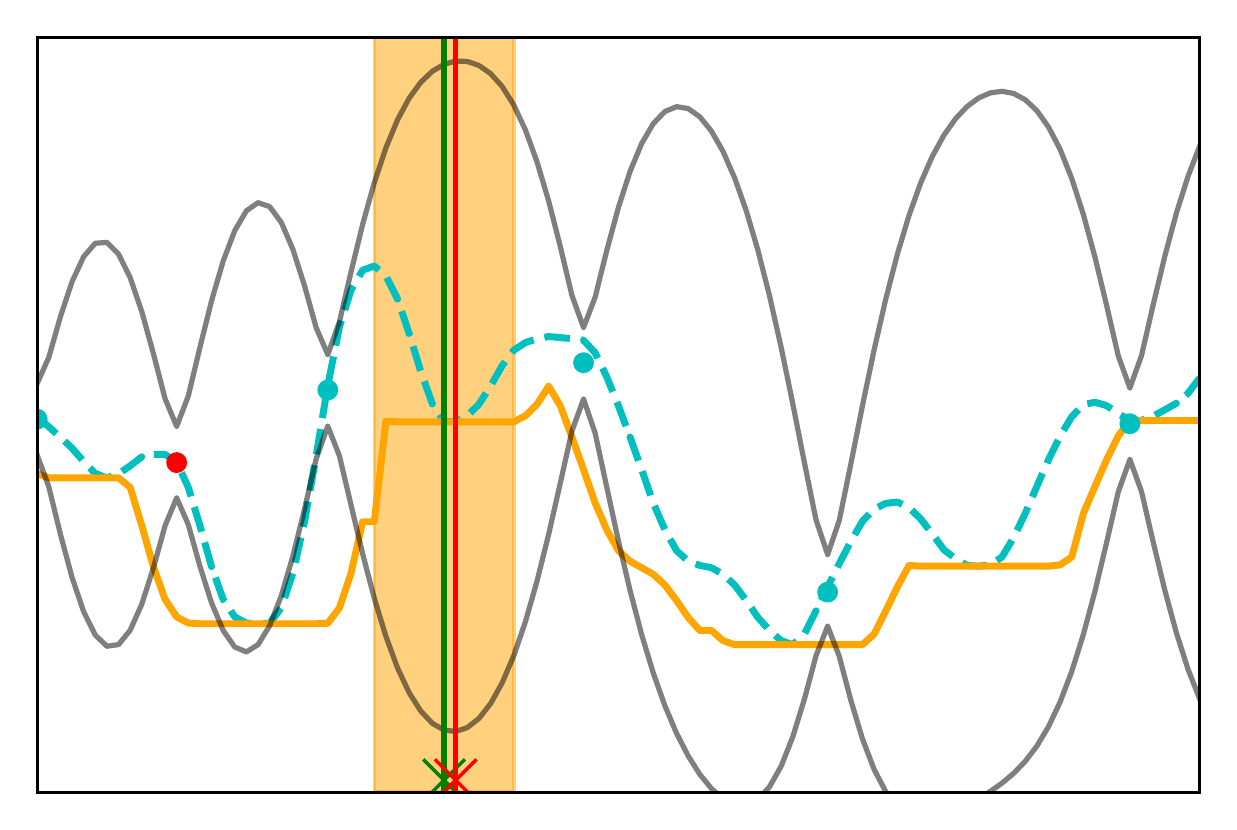}
              \caption{$t=6$}
            \end{subfigure}
            \hspace{0.5cm}
            \begin{subfigure}{.3\textwidth}
              \centering
              \includegraphics[scale=0.35]{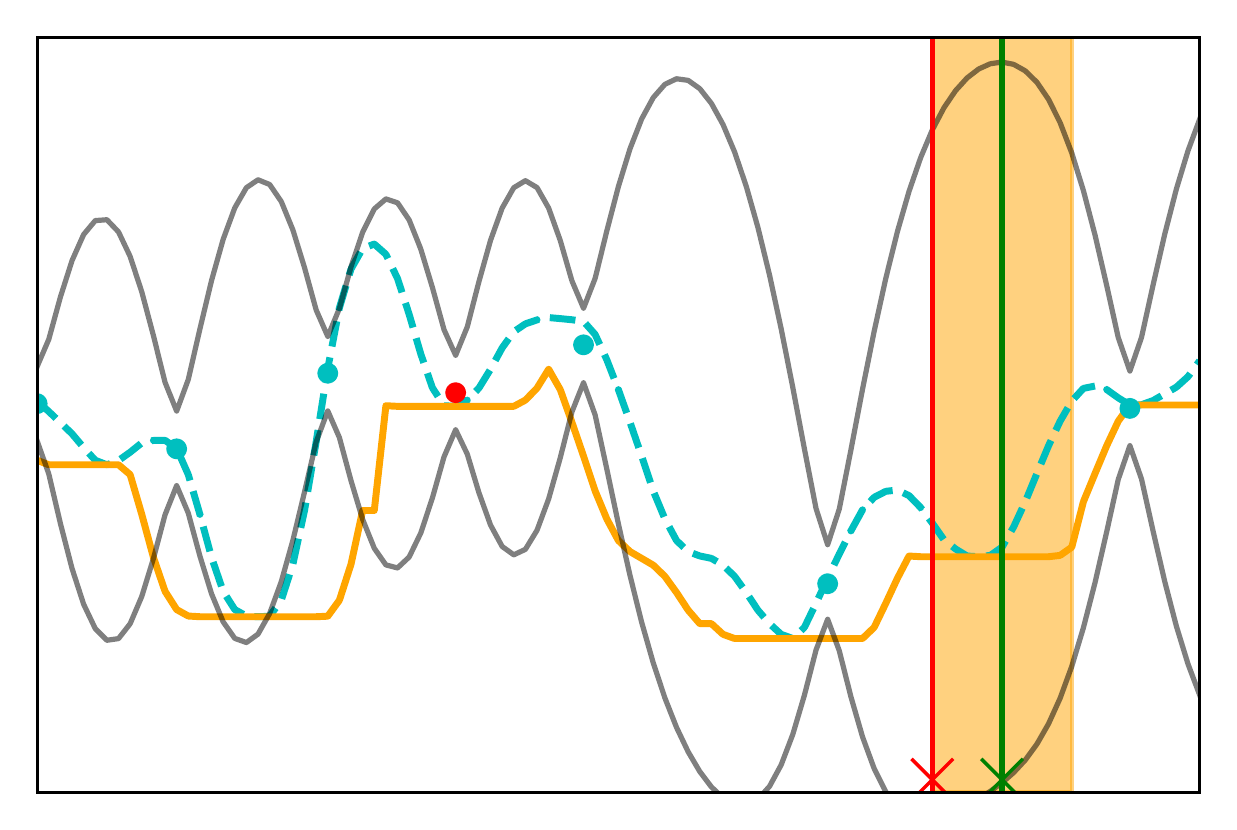}
              \caption{$t=7$}
            \end{subfigure}
            \hspace{0.5cm}
            \begin{subfigure}{.3\textwidth}
              \centering
              \includegraphics[scale=0.35]{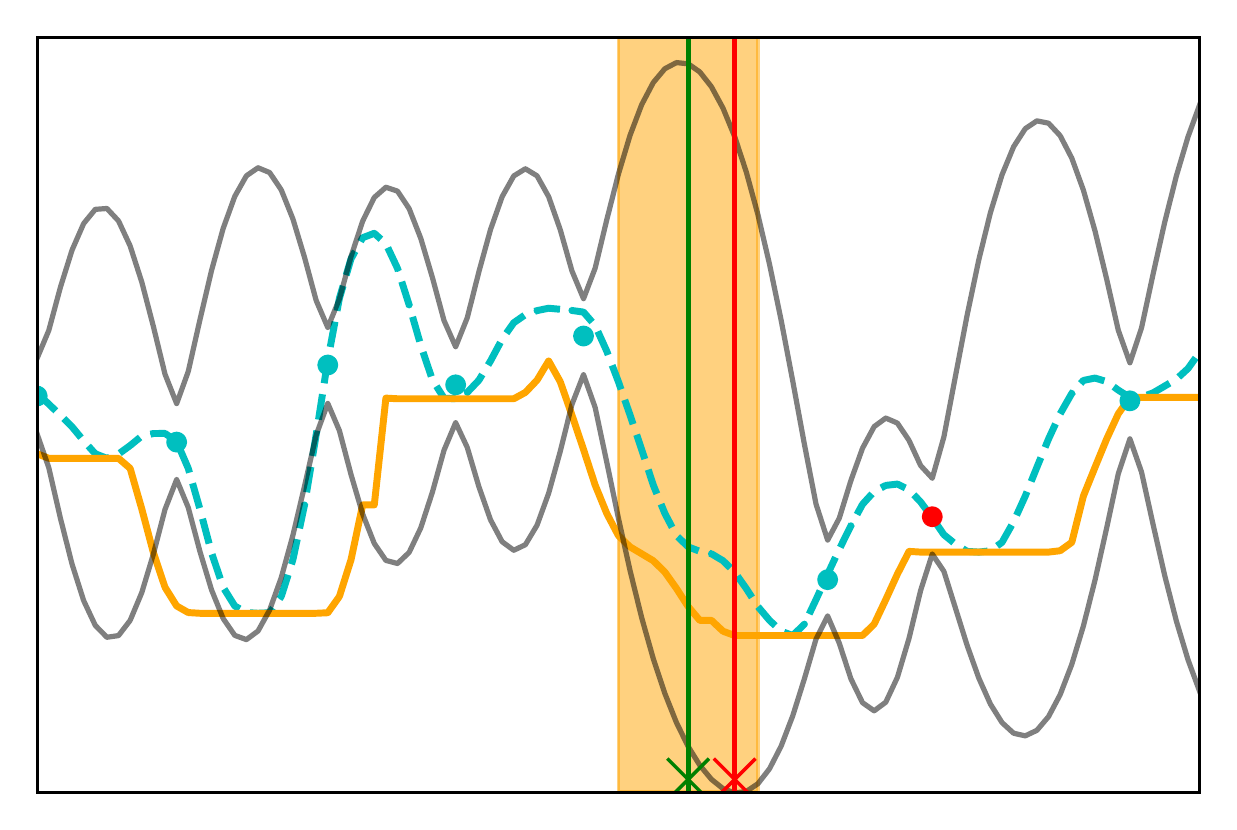}
              \caption{$t=8$}
            \end{subfigure}
            \begin{subfigure}{.3\textwidth}
              \centering
              \includegraphics[scale=0.35]{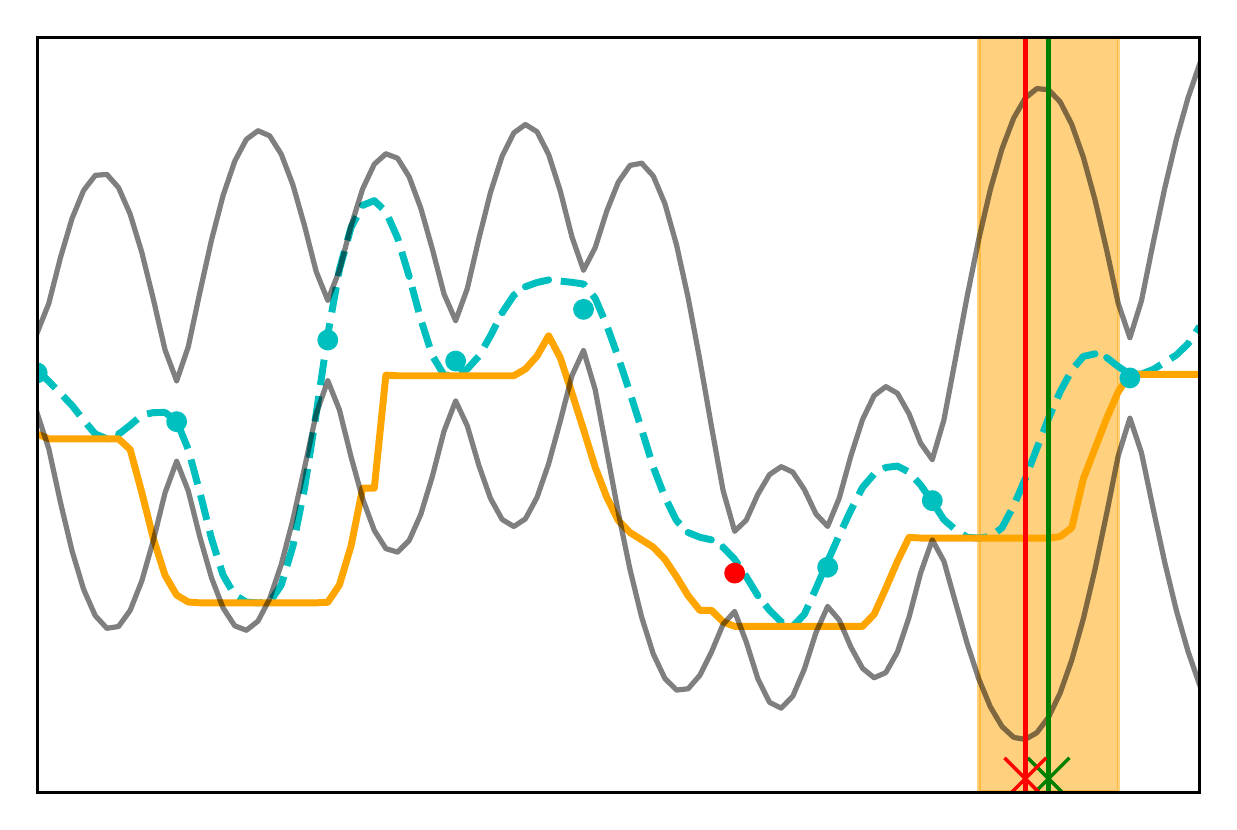}
              \caption{$t=9$}
            \end{subfigure}
            \hspace{0.5cm}
            \begin{subfigure}{.3\textwidth}
              \centering
              \includegraphics[scale=0.35]{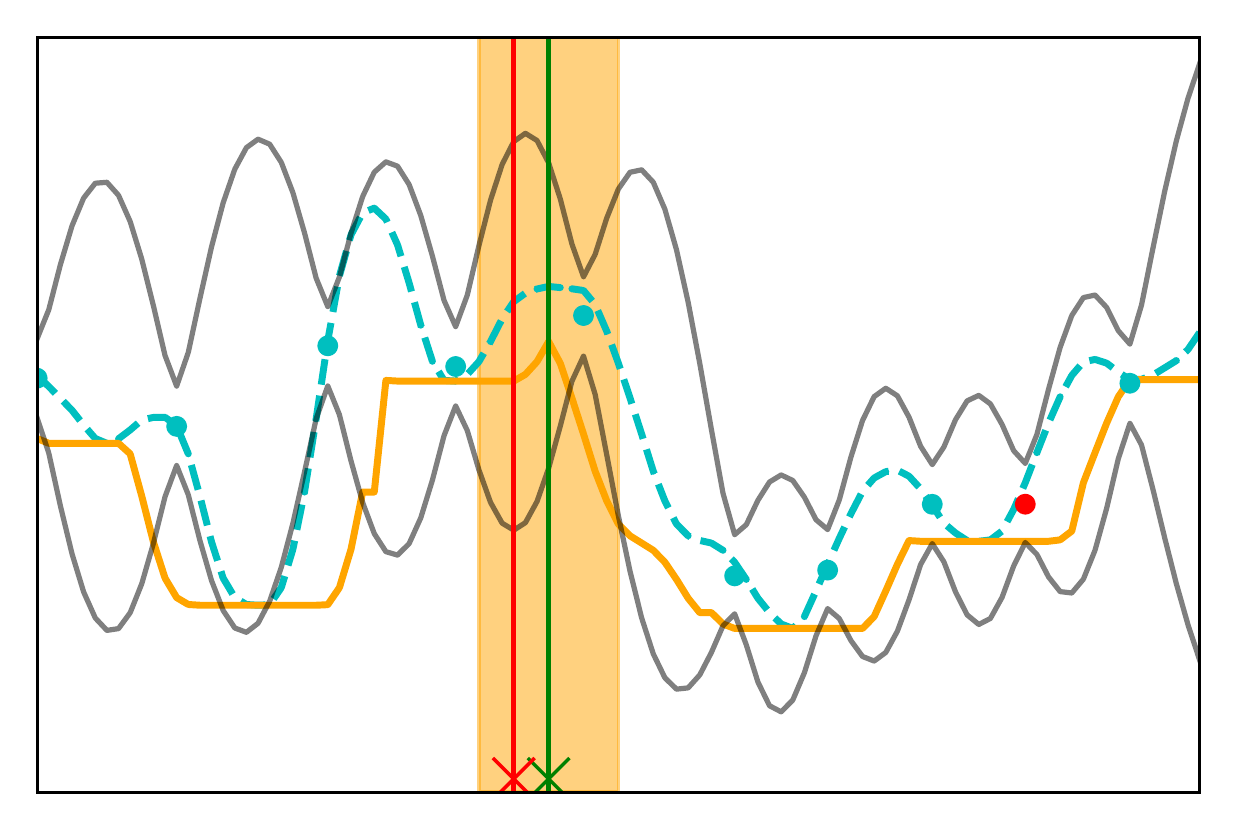}
              \caption{$t=10$}
            \end{subfigure}
            \hspace{0.5cm}
            \begin{subfigure}{.3\textwidth}
              \centering
              \includegraphics[scale=0.35]{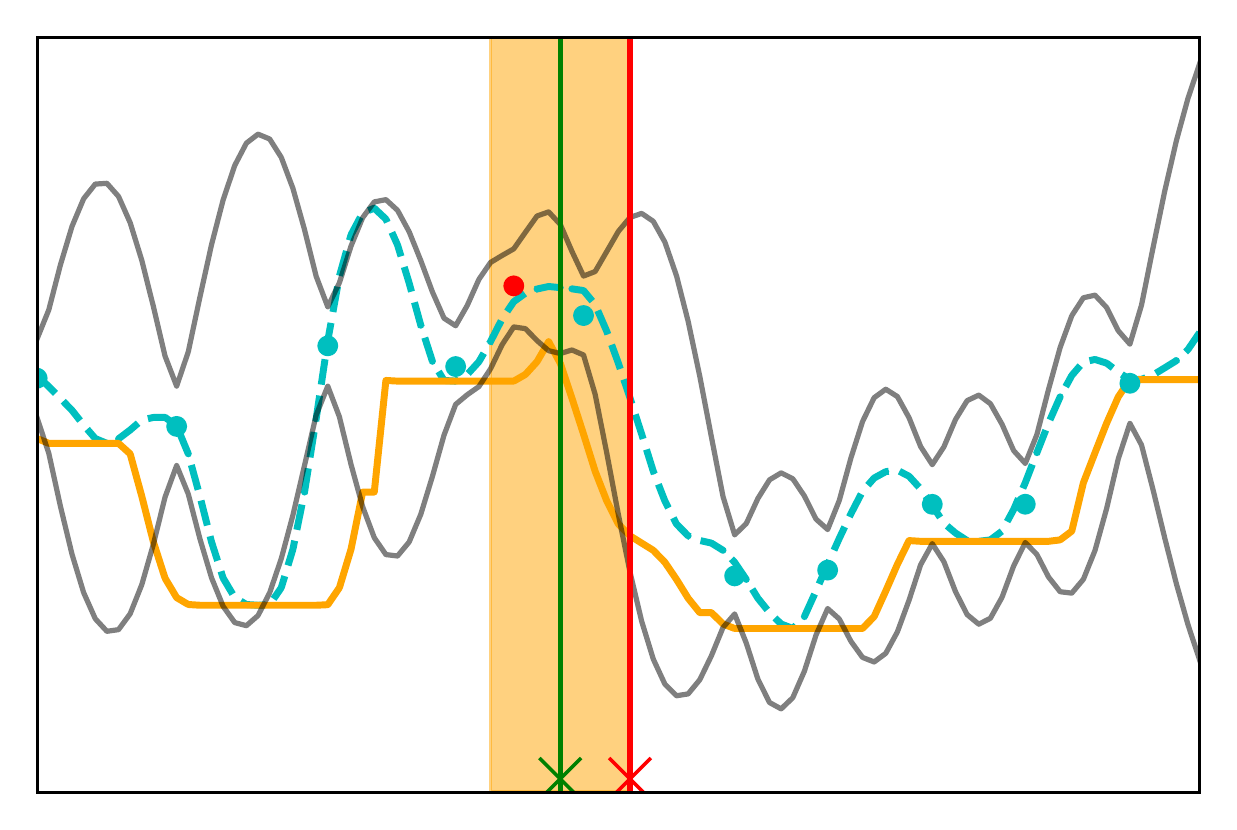}
              \caption{$t=11$}
            \end{subfigure}
            \begin{subfigure}{.3\textwidth}
              \centering
              \includegraphics[scale=0.35]{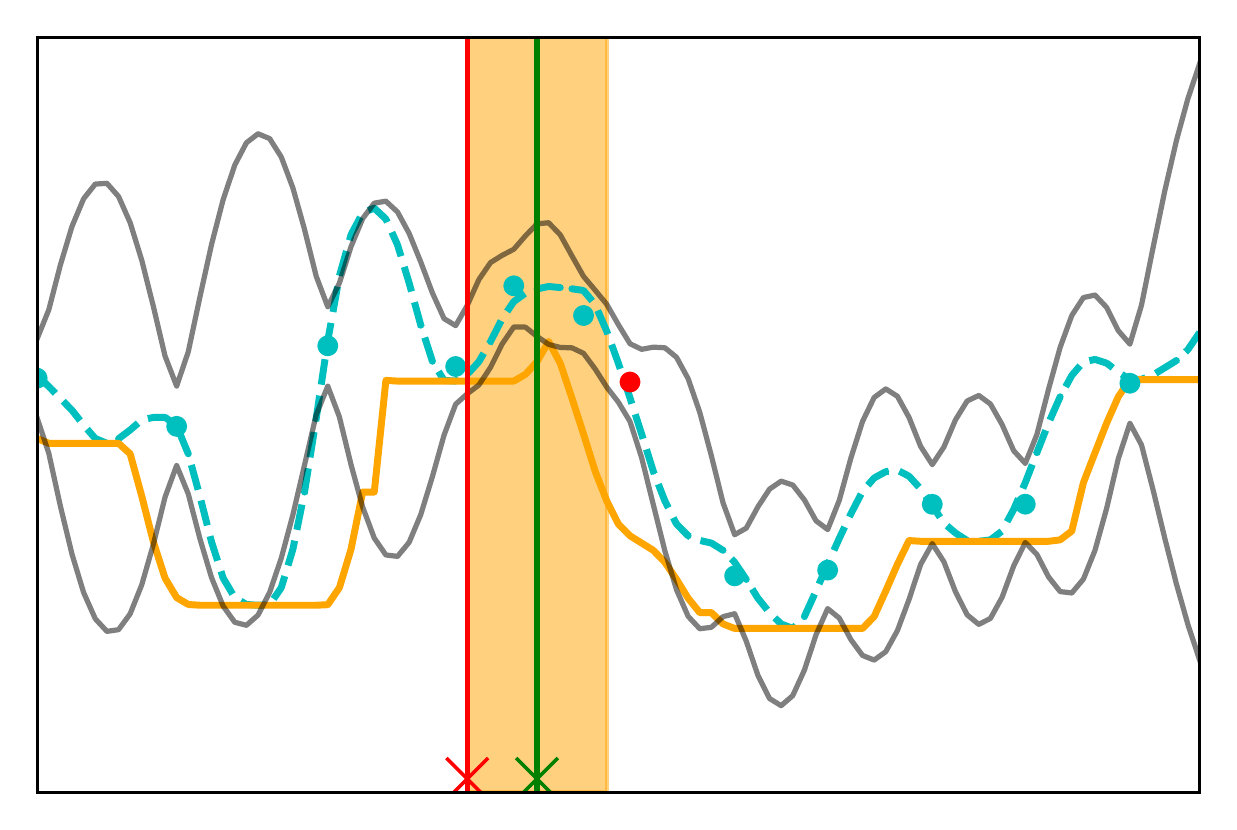}
              \caption{$t=12$}
            \end{subfigure}
            \hspace{0.5cm}
            \begin{subfigure}{.3\textwidth}
              \centering
              \includegraphics[scale=0.35]{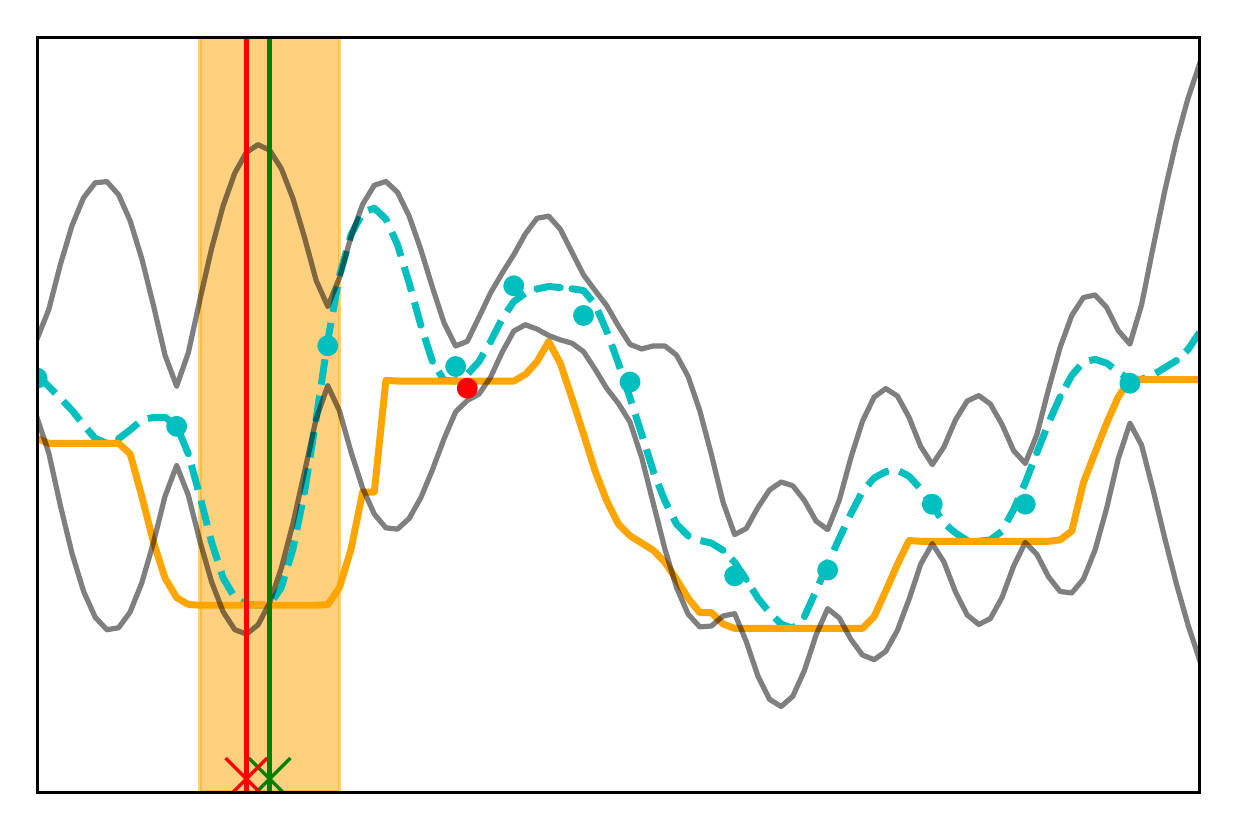}
              \caption{$t=13$}
            \end{subfigure}
            \hspace{0.5cm}
            \begin{subfigure}{.3\textwidth}
              \centering
              \includegraphics[scale=0.35]{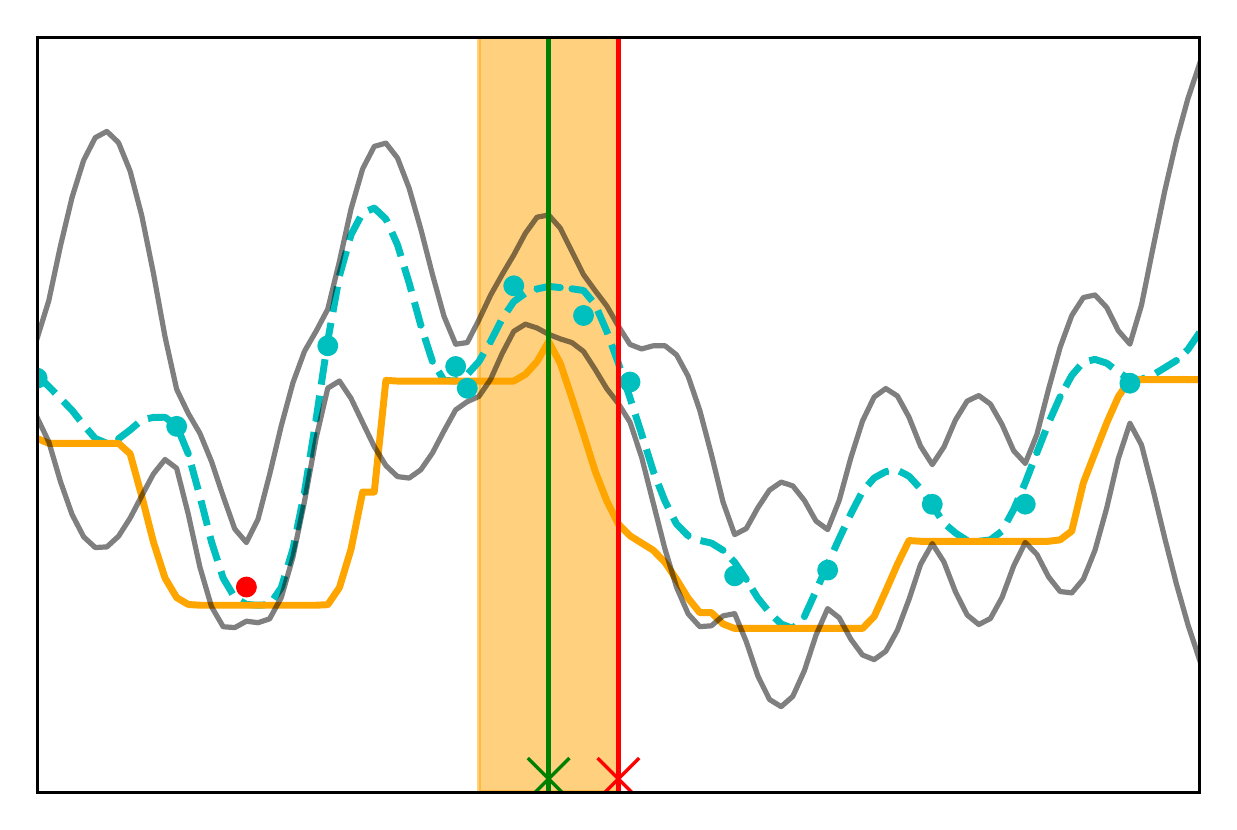}
              \caption{$t=14$}
            \end{subfigure}
    \end{figure}

\section{Proofs of Theoretical Results}

\subsection{Proof of Theorem \ref{thm:upper} (upper bound)}

Recall that $\xtil_t$ is the point computed by \alg~in \eqref{eq:robust_ucb_rule} at time $t$, and that $\vdelta_t$ corresponds to the perturbation obtained in \alg~(Line 3) at time $t$.  In the following, we condition on the event in Lemma \ref{confidence_lemma} holding true, meaning that $\ucb_{t}$ and $\lcb_{t}$ provide valid confidence bounds as per \eqref{eq:conf_bounds}.  As stated in the lemma, this holds with probability at least $1-\xi$.

By the definition of $\epsilon$-instant regret, we have
\begin{align}
r_{\epsilon}(\xtil_t) &= \max_{\x \in D} \min_{\vdelta \in \Delta_{\epsilon}(\x)} f(\x + \vdelta) - \min_{\vdelta \in \Delta_{\epsilon}(x_t)} f(\xtil_t + \vdelta) \\
&\leq \max_{\x \in D} \min_{\vdelta \in \Delta_{\epsilon}(\x)} f(\x + \vdelta) - \min_{\vdelta \in \Delta_{\epsilon}(\xtil_t)} \lcb_{t-1}(\xtil_t + \vdelta) \label{eq:lemma_dep_1}\\
&= \max_{\x \in D} \min_{\vdelta \in \Delta_{\epsilon}(\x)} f(\x + \vdelta)  - \lcb_{t-1}(\xtil_t + \vdelta_t) \label{eq:dxt_def} \\
&\leq   \max_{\x \in D} \min_{\vdelta \in \Delta_{\epsilon}(\x)} \ucb_{t-1}(\x + \vdelta) - \lcb_{t-1}(\xtil_t + \vdelta_t) \label{eq:lemma_dep_2}\\
&= \min_{\vdelta \in \Delta_{\epsilon}(\xtil_t)} \ucb_{t-1}(\xtil_t + \vdelta)- \lcb_{t-1}(\xtil_t + \vdelta_t) \label{eq:xt_def}\\
&\leq \ucb_{t-1}(\xtil_t + \vdelta_t) - \lcb_{t-1}(\xtil_t + \vdelta_t) \label{eq:dxt_2} \\
&= 2 \beta_t^{1/2} \sigma_{t-1}(\xtil_t + \vdelta_t), \label{eq:2beta}
\end{align}
where \eqref{eq:lemma_dep_1} and \eqref{eq:lemma_dep_2} follow from Lemma~\ref{confidence_lemma}, \eqref{eq:dxt_def} follows since $\vdelta_t$ minimizes $\lcb_{t-1}$ by definition,  \eqref{eq:xt_def} follows since $\xtil_t$ maximizes the robust upper confidence bound by definition, \eqref{eq:dxt_2} follows by upper bounding the minimum by the specific choice $\vdelta_t \in \Delta_{\epsilon}(\x_t)$, and \eqref{eq:2beta} follows since the upper and lower confidence bounds are separated by $2 \beta_t^{1/2} \sigma_{t-1}(\cdot)$ according to their definitions in \eqref{eq:ucb_lcb_def}.

In fact, the analysis from \eqref{eq:lemma_dep_1} to \eqref{eq:2beta} shows that the following {\em pessimistic estimate} of $r_{\epsilon}(\xtil_t)$ is upper bounded by $2 \beta_t^{1/2} \sigma_{t-1}(\xtil_t + \vdelta_t)$:
\begin{equation}
\rbar_{\epsilon}(\xtil_t) = \max_{\x \in D} \min_{\vdelta \in \Delta_{\epsilon}(\x)} f(\x + \vdelta) - \min_{\vdelta \in \Delta_{\epsilon}(\xtil_t)} \lcb_{t-1}(\xtil_t + \vdelta). \label{eq:rbar}
\end{equation}
Unlike $r_{\epsilon}(\xtil_t)$, the algorithm has the required knowledge to identify the value of $t \in \{1,\dotsc,T\}$ with the smallest $\rbar_{\epsilon}(\xtil_t)$.  Specifically, the first term on the right-hand side of \eqref{eq:rbar} does not depend on $t$, so the smallest $\rbar_{\epsilon}(\xtil_t)$ is achieved by $\x^{(T)}$ defined in \eqref{eq:final_point}.  Since the minimum is upper bounded by the average, it follows that
\begin{align}
r_{\epsilon}(\x^{(T)}) 
&\le \rbar_{\epsilon}(\x^{(T)}) \\
&\le \frac{1}{T}\sum_{t=1}^T 2 \beta_t^{1/2} \sigma_{t-1}(\xtil_t + \vdelta_t) \label{eq:last_step0} \\
&\le \frac{2\beta_T^{1/2}}{T} \sum_{t=1}^T \sigma_{t-1}(\xtil_t + \vdelta_t), \label{eq:last_step}
\end{align}
where \eqref{eq:last_step0} uses \eqref{eq:2beta}, and \eqref{eq:last_step} uses the monotonicity of $\beta_T$.  Next, we claim that
\begin{equation}
    2\sum_{t=1}^T \sigma_{t-1}(\xtil_t + \vdelta_t) \leq \sqrt{C_1 T \gamma_T}, \label{eq:sum_bound}
\end{equation}
where $C_1 = 8 / \log(1 + \sigma^{-2})$.  In fact, this is a special case of the well-known result \cite[Lemma 5.4]{srinivas2009gaussian},\footnote{More precisely, \cite[Lemma 5.4]{srinivas2009gaussian} alongside an application of the Cauchy-Schwarz inequality as in \cite{srinivas2009gaussian}.} which upper bounds the sum of posterior standard deviations of sampled points in terms of the information gain $\gamma_T$ (recall that \alg~samples at location $\xtil_t + \vdelta_t$).  Combining \eqref{eq:last_step}--\eqref{eq:sum_bound} and re-arranging, we deduce that after $T$ satisfies $\frac{T}{\beta_T \gamma_T} \geq \frac{C_1}{\eta^2}$, the $\epsilon$-instant regret is at most $\eta$, thus completing the proof.

\subsection{Proof of Theorem \ref{thm:lower} (lower bound)}

Our lower bounding analysis builds heavily on that of the non-robust optimization setting with $f \in \Fc_k(B)$ studied in \cite{scarlett2017lower}, but with important differences.  Roughly speaking, the analysis of \cite{scarlett2017lower} is based on the difficulty of finding a very narrow ``bump'' of height $2\eta$ in a function whose values are mostly close to zero.  In the $\epsilon$-stable setting, however, even the points around such a bump will be adversarially perturbed to another point whose function value is nearly zero.  Hence, all points are essentially equally bad.

\begin{figure}
    \begin{centering}
        \includegraphics[width=0.7\columnwidth]{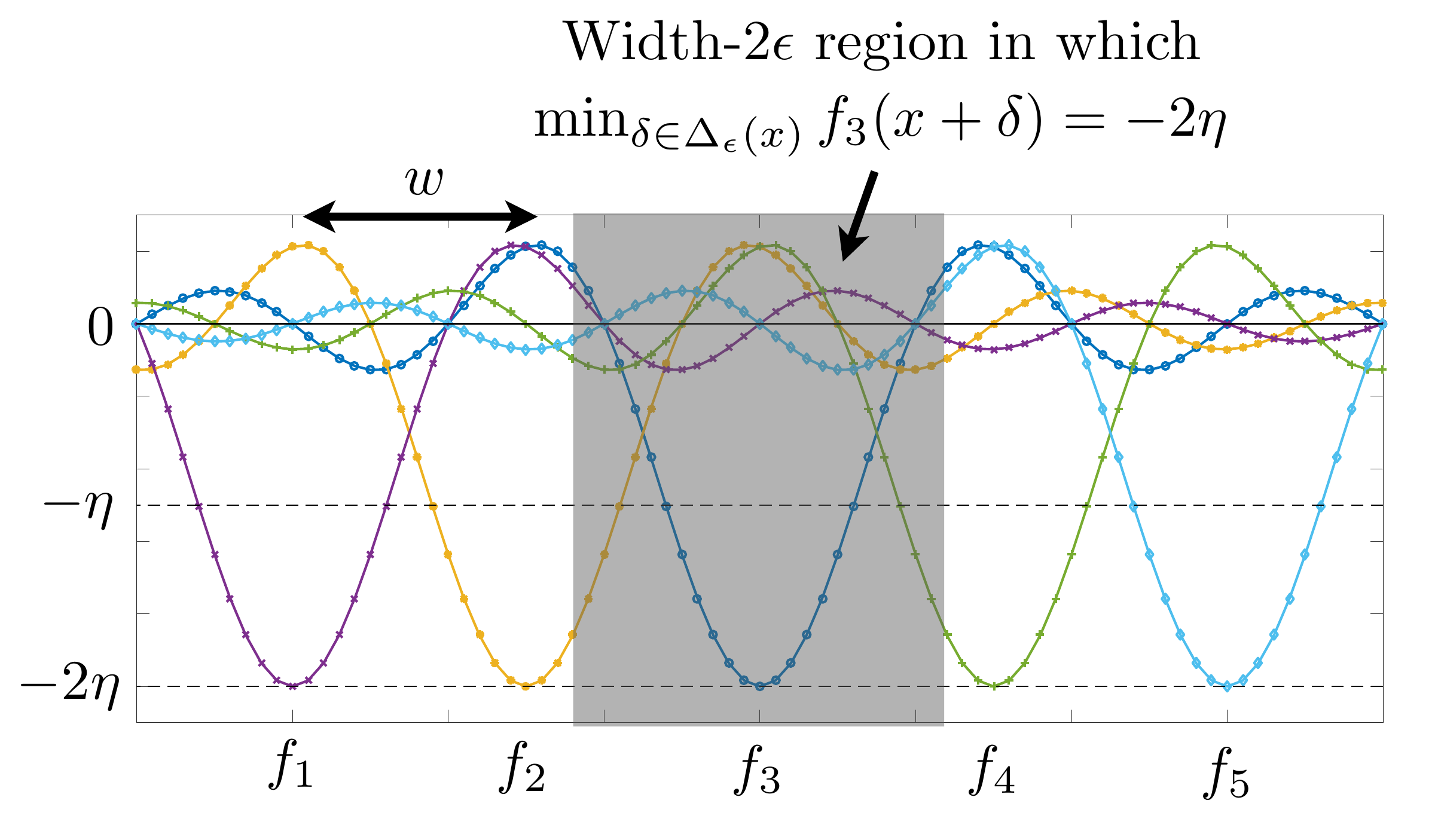}
        \par
    \end{centering}
    
    \caption{Illustration of functions $f_1,\dotsc,f_5$ equal to a common function shifted by various multiples of a given parameter $w$.  In the $\epsilon$-stable setting, there is a wide region (shown in gray for the dark blue curve $f_3$) within which the perturbed function value equals $-2\eta$. \label{fig:func_class}}
\end{figure}

To overcome this challenge, we consider the reverse scenario: Most of the function values are still nearly zero, but there exists a narrow {\em valley} of depth $-2\eta$.   This means that every point within an $\epsilon$-ball around the function minimizer will be perturbed to the point with value $-2\eta$.  Hence, a constant fraction of the volume is still $2\eta$-suboptimal, and it is impossible to avoid this region with high probability unless the time horizon $T$ is sufficiently large.  An illustration is given in Figure \ref{fig:func_class}, with further details below.

We now proceed with the formal proof.

\subsubsection{Preliminaries}

Recall that we are considering an arbitrary given (deterministic) GP optimization algorithm.  More precisely, such an algorithm consists of a sequence of decision functions that return a sampling location $\x_t$ based on $y_1,\dotsc,y_{t-1}$, and an additional decision function that reports the final point $\x^{(T)}$ based on $y_1,\dotsc,y_T$.  The points $\x_1,\dotsc,\x_{t-1}$ (or $\x_1,\dotsc,\x_T$) do not need to be treated as additional inputs to these functions, since $(\x_1,\dotsc,\x_{t-1})$ is a deterministic function of $(y_1,\dotsc,y_{t-1})$.

We first review several useful results and techniques from \cite{scarlett2017lower}:
\begin{itemize}[leftmargin=5ex]
    \item We lower bound the worst-case $\epsilon$-regret within $\Fc_k(B)$ by the $\epsilon$-regret averaged over a suitably-designed finite collection $\{f_1,\dotsc,f_M\} \subset \Fc_k(B)$ of size $M$.
    \item We choose each $f_m(\x)$ to be a shifted version of a common function $g(\x)$ on $\RR^p$.  Specifically, each $f_m(\x)$ is obtained by shifting $g(\x)$ by a different amount, and then cropping to $D = [0,1]^p$.  For our purposes, we require $g(\x)$ to satisfy the following properties:
    \begin{enumerate}
        \item The RKHS norm in $\RR^p$ is bounded, $\|g\|_k \le B$;
        \item We have (i) $g(\x) \in [-2\eta,2\eta]$ with minimum value $g(0) = -2\eta$, and (ii) there is a ``width'' $w$ such that $g(\x) > -\eta$ for all $\|\x\|_{\infty} \ge w$;
        \item There are absolute constants $h_0 > 0$ and $\zeta > 0$ such that $g(\x) = \frac{2\eta}{h_0} h\big(\frac{\x\zeta}{w}\big)$ for some function $h(\z)$ that decays faster than any finite power of $\|\z\|_2$ as $\|\z\|_2 \to \infty$.
    \end{enumerate}
    Letting $g(\x)$ be such a function, we construct the $M$ functions by shifting $g(\x)$ so that each $f_m(\x)$ is centered on a unique point in a uniform grid, with points separated by $w$ in each dimension.  Since $D = [0,1]^p$, one can construct
    \begin{equation}
        M = \Big\lfloor \Big( \frac{1}{w} \Big)^p \Big\rfloor \label{eq:Mw}
    \end{equation}
    such functions.  We will use this construction with $w \ll 1$, so that there is no risk of having $M = 0$, and in fact $M$ can be assumed larger than any desired absolute constant.
    \item It is shown in \cite{scarlett2017lower} that the above properties\footnote{Here $g(\x)$ plays the role of $-g(\x)$ in \cite{scarlett2017lower} due to the discussion at the start of this appendix, but otherwise the construction is identical.} can be achieved with
    \begin{equation}
        M = \Bigg\lfloor \Bigg( \frac{ r\sqrt{\log\frac{B (2\pi l^2)^{p/4} h(0)}{2\eta}} }{\zeta \pi l} \Bigg)^p \Bigg\rfloor \label{eq:M_se}
    \end{equation}
    in the case of the SE kernel, and with 
    \begin{equation}
        M = \Big\lfloor \Big( \frac{B c_3}{\eta} \Big)^{p/\nu} \Big\rfloor \label{eq:M_matern}
    \end{equation}
    in the case of the Mat\'ern kernel, where
    \begin{equation}
        c_3 :=  \Big( \frac{r}{\zeta} \Big)^{\nu} \cdot \Bigg( \frac{ c_2^{-1/2} }{ 2 (8\pi^2)^{(\nu + p/2)/2} } \Bigg),
    \end{equation}
    and where $c_2 > 0$ is an absolute constant.  Note that these values of $M$ amount to choosing $w$ in \eqref{eq:Mw}, and the assumption of sufficiently small $\frac{\eta}{B}$ in the theorem statement ensures that $M \gg 1$ (or equivalently $w \ll 1$) as stated above.
    \item Property 2 above ensures that the ``robust'' function value $\min_{\vdelta \in \Delta_{\epsilon}(\x)} f(\x)$ equals $-2\eta$ for any $\x$ whose $\epsilon$-neighborhood includes the minimizer $\x_{\min}$ of $f$, while being $-\eta$ or higher for any input whose entire $\epsilon$-neighborhood is separated from $\x_{\min}$ by at least $w$.  For $w \ll 1$ and $\epsilon < 0.5,$ a point of the latter type is guaranteed to exist, which implies
    \begin{equation}
        r_{\epsilon}(\x) \ge \eta \label{eq:r_lb}
    \end{equation}
    for any $\x$ whose $\epsilon$-neighborhood includes $\x_{\min}$.
\end{itemize}

In addition, we introduce the following notation, also used in \cite{scarlett2017lower}:
\begin{itemize}
    \item The probability density function of the output sequence $\y = (y_1,\dotsc,y_T)$ when the underlying function is $f_m$ is denoted by $P_m(\y)$.  We also define $f_0(\x) = 0$ to be the zero function, and define $P_0(\y)$ analogously for the case that the optimization algorithm is run on $f_0$.  Expectations and probabilities (with respect to the noisy observations) are similarly written as $\EE_m$, $\PP_m$, $\EE_0$, and $\PP_0$ when the underlying function is $f_m$ or $f_0$.  On the other hand, in the absence of a subscript, $\EE[\cdot]$ and $\PP[\cdot]$ are taken with respect to the noisy observations {\em and} the random function $f$ drawn uniformly from $\{f_1,\dotsc,f_M\}$ (recall that we are lower bounding the worst case by this average). 
    \item Let $\{\Rc_m\}_{m=1}^M$ be a partition of the domain into $M$ regions according the above-mentioned uniform grid, with $f_m$ taking its minimum value of $-2\eta$ in the centre of $\Rc_m$.  Moreover, let $j_t$ be the index at time $t$ such that $\x_t$ falls into $\Rc_{j_t}$; this can be thought of as a quantization of $\x_t$.
    % \item Let $P_m(y_t|\y_{t-1})$ denote the distribution of $y_t$ given all the observations up to time $t-1$, denoted by $\y_{t-1} = (y_1,\dotsc,y_{t-1})$ (and $\y_0 = \emptyset$), in the case that $M=m$.
    \item Define the maximum (absolute) function value within a given region $\Rc_j$ as
    \begin{equation}
    \vbar_m^j := \max_{\x \in \Rc_j} |f_m(\x)|, \label{eq:vbar}
    \end{equation}
    and the maximum KL divergence to $P_0$ within the region as
    \begin{equation}
    \Dbar_m^j := \max_{\x \in \Rc_j} D( P_0(\cdot|\x) \| P_m(\cdot|\x) ), \label{eq:Dbar}
    \end{equation}
    where $P_m(y|\x)$ is the distribution of an observation $y$ for a given selected point $\x$ under the function $f_m$, and similarly for $P_0(y|\x)$.
    \item Let $N_j \in \{0,\dotsc,T\}$ be a random variable representing the number of points from $\Rc_j$ that are selected throughout the $T$ rounds.
\end{itemize}

Next, we present several useful lemmas.  The following well-known change-of-measure result, which can be viewed as a form of Le Cam's method, has been used extensively in both discrete and continuous bandit problems.

\begin{lemma} \emph{\cite[p.~27]{auer1995gambling}} \label{lem:auer}
    For any function $a(\y)$ taking values in a bounded range $[0,A]$, we have
    \begin{align}
    \big| \EE_m[a(\y)] - \EE_0[a(\y)]\big| 
        &\le A\, d_{\rm TV}(P_0, P_m) \label{eq:auer_bound} \\
        &\le A\, \sqrt{  D(P_0 \| P_m) }, \label{eq:auer_bound2}
    \end{align}
    where $d_{\rm TV}(P_0,P_m) = \frac{1}{2} \int_{\RR^T} |P_0(\y) - P_m(\y)| \,d\y$ is the total variation distance.
\end{lemma}

We briefly remark on some slight differences here compared to \cite[p.~27]{auer1995gambling}.  There, only $\EE_m[a(\y)] - \EE_0[a(\y)]$ is upper bounded in terms of $d_{\rm TV}(P_0,P_m)$, but one easily obtains the same upper bound on $\EE_0[a(\y)] - \EE_m[a(\y)]$ by interchanging the roles of $P_0$ and $P_m$.  The step \eqref{eq:auer_bound2} follows from Pinsker's inequality, $d_{\rm TV}(P_0, P_m) \le \sqrt{\frac{D(P_0 \| P_m)}{2}}$, and by upper bounding $\frac{1}{\sqrt 2} \le 1$ to ease the notation.

The following result simplifies the divergence term in \eqref{eq:auer_bound2}.

\begin{lemma} \label{lem:div_bound}
    {\em \cite[Eq.~(44)]{scarlett2017lower}}
    Under the preceding definitions, we have
    \begin{equation}
    D(P_0 \| P_m) \le \sum_{j=1}^M \EE_0[N_j]\Dbar_m^j. \label{eq:div_bound}
    \end{equation}
\end{lemma}

The following well-known property gives a formula for the KL divergence between two Gaussians.

\begin{lemma}
    {\em \cite[Eq.~(36)]{scarlett2017lower}}
    For $P_1$ and $P_2$ being Gaussian with means $(\mu_1,\mu_2)$ and a common variance $\sigma^2$, we have
    \begin{equation}
    D(P_1 \| P_2) = \frac{ (\mu_1 - \mu_2)^2 }{ 2\sigma^2 }. \label{eq:Gaussian_div}
    \end{equation}  
\end{lemma}

Finally, we have the following technical result regarding the ``needle-in-haystack'' type function constructed above.

\begin{lemma} \label{lem:vbar_sums}
    {\em \cite[Lemma 7]{scarlett2017lower}}
    The functions $\{f_m\}_{m=1}^M$ corresponding to \eqref{eq:M_se}--\eqref{eq:M_matern} are such that the quantities $\vbar_m^j$ satisfy $\sum_{m=1}^M (\vbar_m^j)^2 = O(\eta^2)$ for all $j$.
\end{lemma}

\subsubsection{Analysis of the average $\epsilon$-stable regret}

Let $\Jbad(m)$ be the set of $j$ such that all $\x \in \Rc_j$ yield $\min_{\vdelta \in \Delta_{\epsilon}(\x)} f(\x + \vdelta) = -2\eta$ when the true function is $f_m$, and define $\Rbad(m) = \cup_{j \in \Jbad(m)} \Rc_j$.  By the $\epsilon$-regret lower bound in \eqref{eq:r_lb}, we have
\begin{align}
    \EE_m[r_{\epsilon}(\x^{(T)})] 
        &\ge \eta \PP_{m}[ \x^{(T)} \in \Rbad(m) ] \\
        &\ge \eta \bigg( \PP_{0}[ \x^{(T)} \in \Rbad(m) ] - \sqrt{D(P_0 \| P_m)} \bigg) \label{eq:rT_step2} \\
        &\ge \eta \bigg( \PP_{0}[ \x^{(T)} \in \Rbad(m) ] - \sqrt{\sum_{j=1}^M \EE_0[N_j]\Dbar_m^j} \bigg), \label{eq:rT_step3}
\end{align}
where \eqref{eq:rT_step2} follows from Lemma \ref{lem:auer} with $a(\y) = \bone\{ \x^{(T)} \in \Rbad(m)\}$ and $A = 1$ (recall that $\x^{(T)}$ is a function of $\y = (y_1,\dotsc,y_T)$), and \eqref{eq:rT_step3} follows from Lemma \ref{lem:div_bound}.  Averaging over $m$ uniform on $\{1,\dotsc,M\}$, we obtain
\begin{equation}
    \EE[r_{\epsilon}(\x^{(T)})] \ge \eta \bigg( \frac{1}{M} \sum_{m=1}^M \PP_{0}[ \x^{(T)} \in \Rbad(m) ] -  \frac{1}{M} \sum_{m=1}^M\sqrt{\sum_{j=1}^M \EE_0[N_j]\Dbar_m^j} \bigg). \label{eq:two_terms}
\end{equation}
We proceed by bounding the two terms separately.
\begin{itemize}[leftmargin=5ex]
    \item We first claim that
    \begin{equation}
        \frac{1}{M} \sum_{m=1}^M \PP_{0}[ \x^{(T)} \in \Rbad(m) ] \ge C_1 \label{eq:claim1}
    \end{equation}
    for some $C_1 > 0$.  To show this, it suffices to prove that any given $\x^{(T)} \in D$ is in at least a constant fraction of the $\Rbad(m)$ regions, of which there are $M$.  This follows from the fact that the $\epsilon$-ball centered at $\x_{m,\min} = \argmin_{\x \in D} f_m(\x)$ takes up a constant fraction of the volume of $D$, where the constant depends on both the stability parameter $\epsilon$ and the dimension $p$.   A small caveat is that because the definition of $\Rbad$ insists that the {\em every} point in the region $\Rc_j$ is within distance $\epsilon$ of $\x_{m,\min}$, the left-hand side of \eqref{eq:claim1} may be slightly below the relevant ratio of volumes above.  However, since Theorem \ref{thm:lower} assumes that $\frac{\epsilon}{B}$ is sufficiently small, the choices of $M$ in \eqref{eq:M_se} and \eqref{eq:M_matern} ensure that $M$ is sufficiently large for this ``quantization'' effect to be negligible.
    \item For the second term in \eqref{eq:two_terms}, we claim that
    \begin{equation}
        \frac{1}{M} \sum_{m=1}^M\sqrt{\sum_{j=1}^M \EE_0[N_j]\Dbar_m^j} \le C_2 \frac{\eta}{\sigma} \sqrt{\frac{T}{M}} \label{eq:claim2}
    \end{equation}
    for some $C_2 > 0$.  To see this, we write
    \begin{align}
        & \frac{1}{M}\sum_{m=1}^M \sqrt{ \sum_{j=1}^M \EE_0[N_{j}]\Dbar_m^{j} } \nonumber \\
        & \qquad = O\bigg( \frac{1}{\sigma} \bigg)\cdot\frac{1}{M}\sum_{m=1}^M  \sqrt{ \sum_{j=1}^M \EE_0[N_{j}](\vbar_m^{j})^2 } \label{eq:second_term_2} \\
        &\qquad\le O\bigg(\frac{1}{\sigma}\bigg)\cdot\sqrt{ \frac{1}{M}\sum_{m=1}^M \sum_{j=1}^M \EE_0[N_{j}](\vbar_m^{j})^2 } \label{eq:second_term_4} \\
        &\qquad =O\bigg(\frac{1}{\sigma}\bigg)\cdot\sqrt{ \frac{1}{M} \sum_{j=1}^M \EE_0[N_{j}]\bigg( \sum_{m=1}^M (\vbar_m^{j})^2\bigg) } \label{eq:second_term_5} \\
        &\qquad =O\bigg(\frac{\eta}{\sqrt{M} \sigma}\bigg)\cdot\sqrt{ \sum_{j=1}^M \EE_0[N_{j}] } \label{eq:second_term_6} \\
        &\qquad = O\bigg(\frac{\sqrt{T}\eta}{\sqrt{M} \sigma}\bigg),\label{eq:second_term_7}
    \end{align} 
    where \eqref{eq:second_term_2} follows since the divergence $D( P_0(\cdot|\x) \| P_m(\cdot|\x) )$ associated with a point $\x$ having value $v(\x)$ is $\frac{v(\x)^2}{2\sigma^2}$ (\emph{cf.}, \eqref{eq:Gaussian_div}), \eqref{eq:second_term_4} follows from Jensen's inequality, \eqref{eq:second_term_6} follows from Lemma \ref{lem:vbar_sums}, and \eqref{eq:second_term_7} follows from $\sum_{j} N_{j} = T$.
\end{itemize}
Substituting \eqref{eq:claim1} and \eqref{eq:claim2} into \eqref{eq:two_terms}, we obtain
\begin{equation}
    \EE[r_{\epsilon}(\x^{(T)})] \ge \eta\Big( C_1 - C_2 \frac{\eta}{\sigma} \sqrt{\frac{T}{M}}\Big),
\end{equation}
which implies that the regret is lower bounded by $\Omega(\eta)$ unless $T = \Omega\big( \frac{M \sigma^2}{ \eta^2 }\big)$.  Substituting $M$ from \eqref{eq:M_se} and \eqref{eq:M_matern}, we deduce that the conditions on $T$ in the theorem statement are necessary to achieve average regret $\EE[r_{\epsilon}(\x^{(T)})] = O(\eta)$ with a sufficiently small implied constant.

\subsubsection{From average to high-probability regret}

Recall that we are considering functions whose values lie in the range $[-2\eta,2\eta]$, implying that $r_{\epsilon}(\x^{(T)}) \le 4\eta$.  Letting $T_{\eta}$ be the lower bound on $T$ derived above for achieving average regret $O(\eta)$ (i.e., we have $\EE[ r_{\epsilon}^{(T_\eta)} ] = \Omega(\eta)$), it follows from the reverse Markov inequality (i.e., Markov's inequality applied to the random variable $4\eta - r_{\epsilon}^{(T_\eta)}$) that
\begin{equation}
    \PP[r_{\epsilon}(\x^{(T_\eta)}) \ge c\eta] \ge \frac{\Omega(\eta) - c\eta}{4\eta - c\eta}
\end{equation}
for any $c > 0$ sufficiently small for the numerator and denominator to be positive.  The right-hand side is lower bounded by a constant for any such $c$, implying that the probability of achieving $\epsilon$-regret at most $c\eta$ cannot be arbitrarily close to one.  By renaming $c\eta$ as $\eta'$, it follows that in order to achieve some target $\epsilon$-stable regret $\eta'$ with probability sufficiently close to one, a lower bound of the same form as the average regret bound holds.  In other words, the conditions on $T$ in the theorem statement remain necessary also for the high-probability regret.

We emphasize that Theorem \ref{thm:lower} concerns the high-probability regret when ``high probability'' means {\em sufficiently close to one} as a function of $\epsilon$, $p$, and the kernel parameters (but still constant with respect to $T$ and $\eta$).  We do not claim a lower bound under any particular {\em given} success probability (e.g., $\eta$-optimality with probability at least $\frac{3}{4}$).

\section{Details on Variations from Section \ref{sec:variations}}

We claim that the \alg~variations and theoretical results outlined in Section \ref{sec:variations} are in fact {special cases} of Algorithm \ref{alg:stable_opt} and Theorem \ref{thm:upper}, despite being seemingly quite different.  The idea behind this claim is that Algorithm \ref{alg:stable_opt} and Theorem \ref{thm:upper} allow for the ``distance'' function $d(\cdot,\cdot)$ to be {completely arbitrary}, so we may choose it in rather creative/unconventional ways.

In more detail, we have the following:
\begin{itemize}[leftmargin=5ex]
    \item For the unknown parameter setting $\max_{\x \in D} \min_{\vtheta \in \Theta} f(\x, \vtheta)$, we replace  $\x$ in the original setting by the concatenated input $(\x,\vtheta)$, and set 
    \begin{equation}   
        d( (\x,\vtheta), (\x',\vtheta') ) = \|\x - \x'\|_2.
    \end{equation}
    If we then set $\epsilon = 0$, we find that the input $\x$ experiences no perturbation, whereas $\vtheta$ may be perturbed arbitrarily, thereby reducing \eqref{eq:eps_stable_input} to $\max_{\x \in D} \min_{\vtheta \in \Theta} f(\x, \vtheta)$ as desired.  
    \item For the robust estimation setting, we again use the concatenated input $(\x,\vtheta)$.  To avoid overloading notation, we let $d_0(\vtheta,\vtheta')$ denote the distance function (applied to $\vtheta$ alone) adopted for this case in Section \ref{sec:variations}.  We set
    \begin{equation}   
            d((\x,\vtheta), (\x',\vtheta')) = 
        \begin{cases}
        d_{0}(\vtheta,\vtheta') & \x = \x' \\
        \infty  &  \x \ne \x'.
        \end{cases}
    \end{equation}
    Due to the second case, the input $\x$ experiences no perturbation, since doing so would violate the distance constraint of $\epsilon$.  We are then left with $\x = \x'$ and $d_{0}(\vtheta,\vtheta') \le \epsilon$, as required.
%    for some constant $\alpha>0$. We apply Theorem~\ref{thm:upper} with the given $\epsilon$, $\x$ replaced by $(\x, \bar{\vtheta})$ and $$ d( (\x,\bar{\vtheta}), (\x',\bar{\vtheta}') ) = d_2 (\x,\x') + d_1(\bar{\vtheta}, \bar{\vtheta}').$$
%    We observe that input $\x$ experiences no perturbation as any perturbation to $\x$ will immediately violate the given budget $\epsilon$.
    \item For the grouped setting $\max_{G \in \mathcal{G}} \min_{\x \in G} f(\x)$, we adopt the function
    \begin{equation}    
        d(\x,\x') = \mathbf{1}\{ \x\text{ and }\x'\text{ are in different groups} \},
    \end{equation}
    and set $\epsilon = 0$.  Considering the formulation in \eqref{eq:eps_stable_input}, we find that any two inputs $\x$ and $\x'$ yield the same $\epsilon$-stable objective function, and hence, reporting a point $\x$ is equivalent to reporting its group $G$.  As a result, \eqref{eq:eps_stable_input} reduces to the desired formulation $\max_{G \in \mathcal{G}} \min_{\x \in G} f(\x)$.
\end{itemize}
The variations of \alg~described in \eqref{eq:variation1}--\eqref{eq:variation3}, as well as the corresponding theoretical results outlined in Section \ref{sec:variations}, follow immediately by substituting the respective choices of $d(\cdot,\cdot)$ and $\epsilon$ above into Algorithm \ref{alg:stable_opt} and Theorem \ref{thm:upper}.  It should be noted that in the first two examples, the definition of $\gamma_t$ in \eqref{eq:gamma_def} is modified to take the maximum over not only $\x_1, \cdots, \x_t$, but also $\vtheta_1, \cdots, \vtheta_t$.

\section{Lake Data Experiment}

We consider an application regarding environmental monitoring of inland waters, using a data set containing 2024 in situ measurements of chlorophyll concentration within a vertical transect plane, collected by an autonomous surface vessel in Lake Z\"{u}rich. This data set was considered in previous works such as \cite{bogunovic2016truncated,gotovos2013active} to detect regions of high concentration.  In these works, the goal was to locate all regions whose concentration exceeds a pre-defined threshold.

\begin{figure}[h!]
    \begin{subfigure}{.33\textwidth}
        \centering
        \includegraphics[scale=0.35]{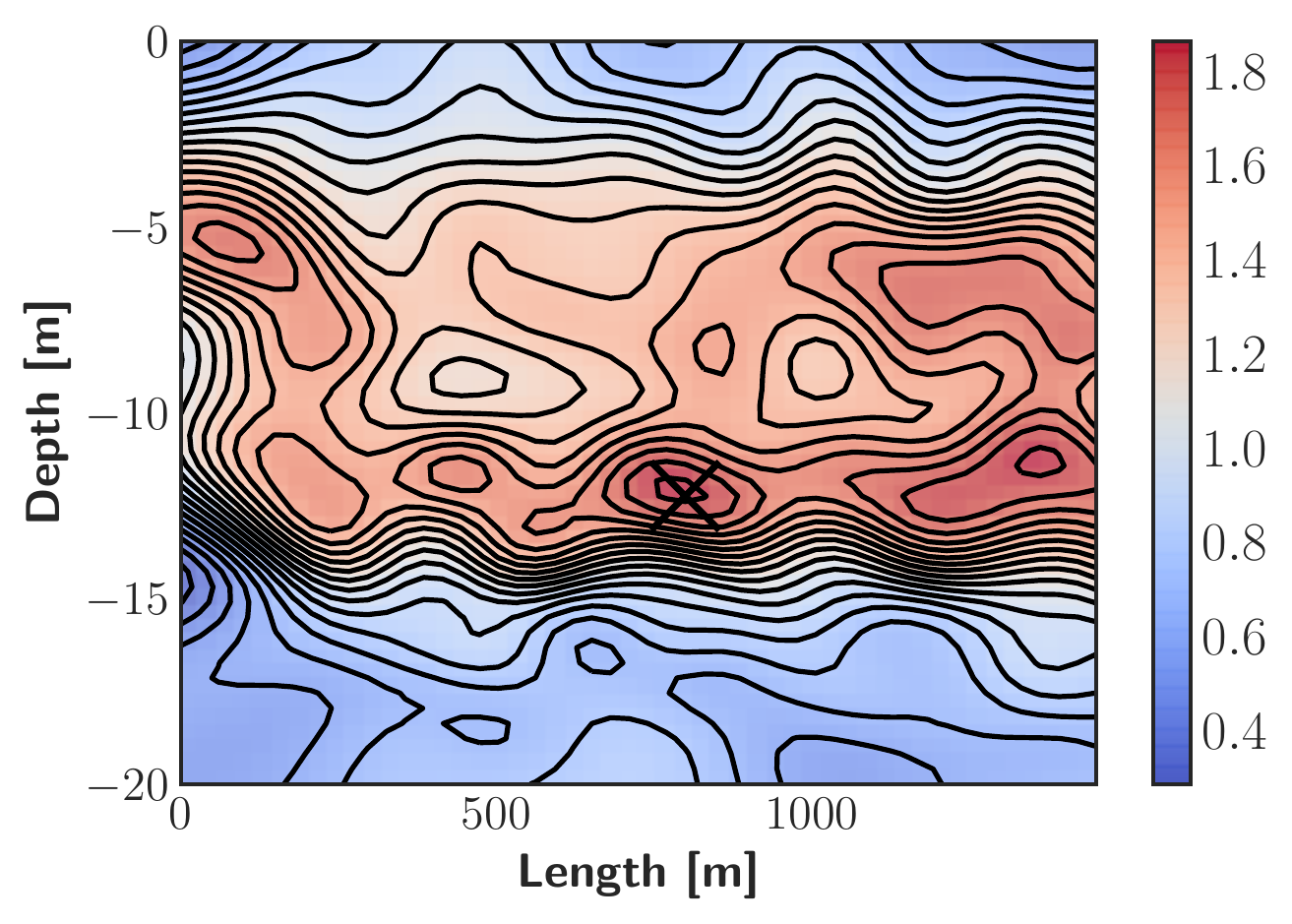}
        \caption{Chlorophyll concentration}
    \end{subfigure}
    \begin{subfigure}{.33\textwidth}
        \centering
        \includegraphics[scale=0.35]{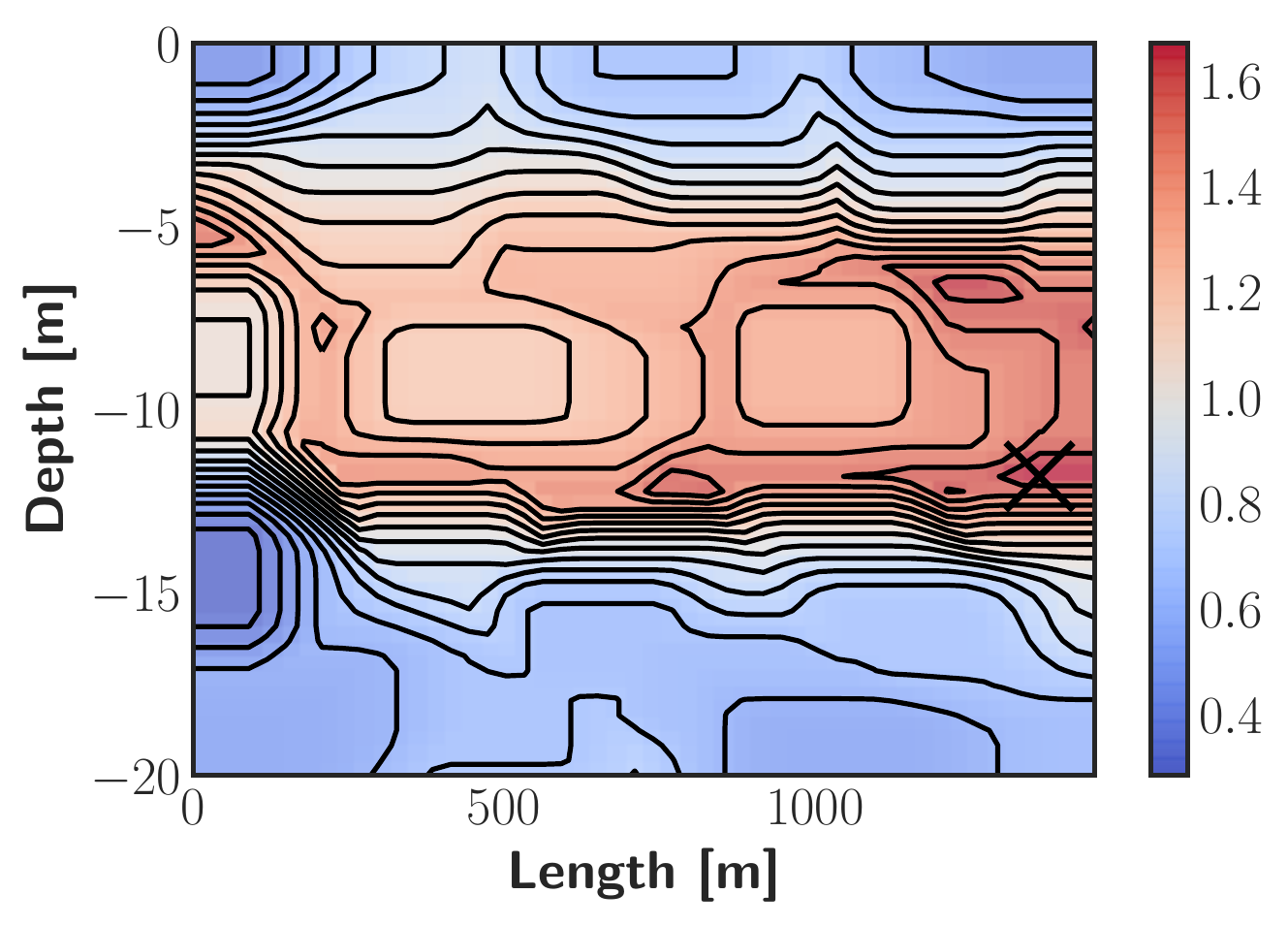}
        \caption{Robust objective}
    \end{subfigure}
    \begin{subfigure}{.33\textwidth}
        \centering
        \includegraphics[scale=0.35]{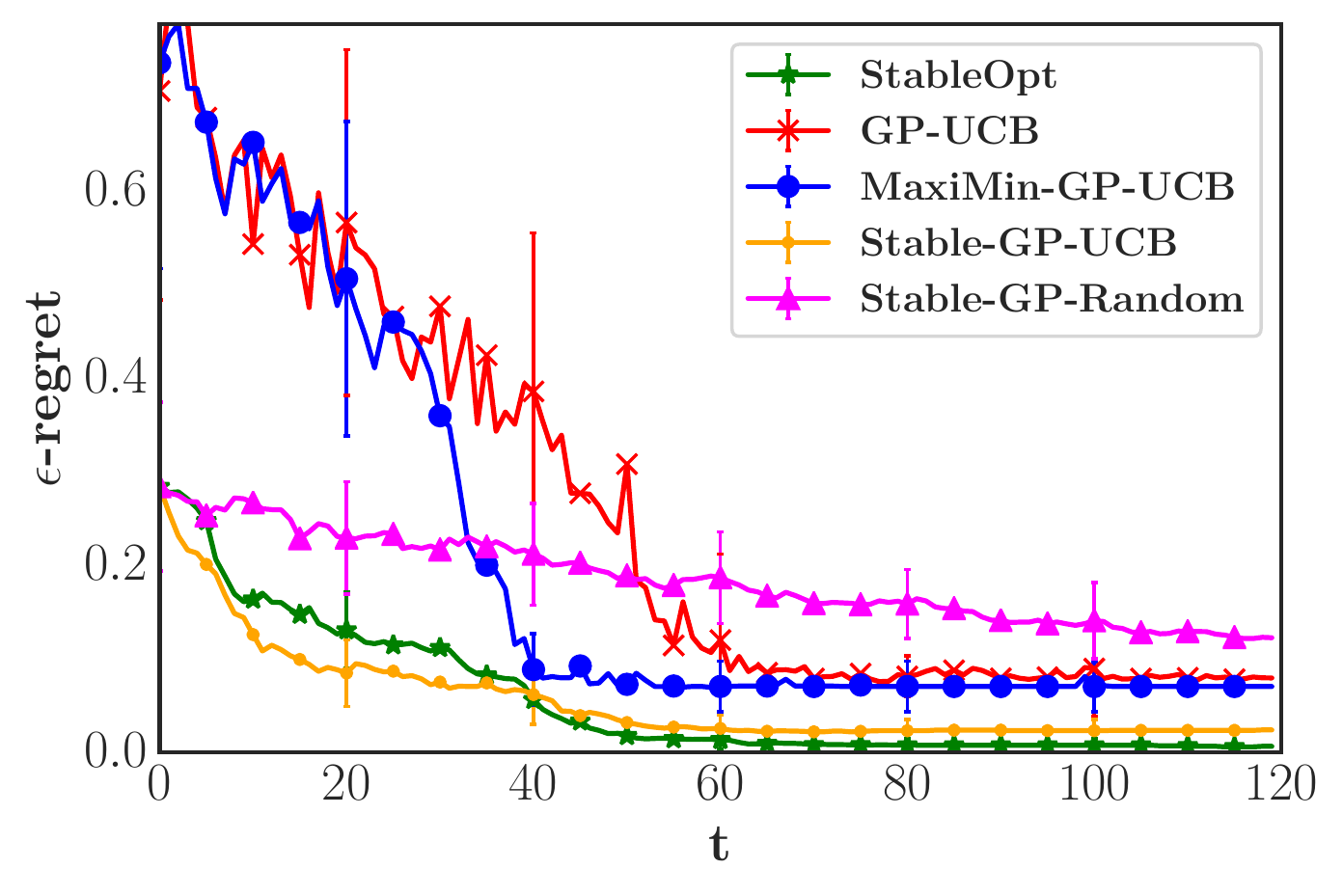}
        \caption{$\epsilon$-regret}
    \end{subfigure}
    \caption{Experiment on the Z\"urich lake dataset; In the later rounds \alg~is the only method that reports a near-optimal $\epsilon$-stable point.}
    \label{fig:lake}
\end{figure}

Here we consider a different goal: We seek to locate a region of a given size such that the concentration throughout the region is as high as possible (in the max-min sense).  This is of interest in cases where high concentration only becomes relevant when it is spread across a sufficiently wide area.  We consider rectangular regions with different pre-specified lengths in each dimension:
\begin{equation}
\Delta_{\epsilon_D, \epsilon_L}(\x) = \{ \x' - \x  \,:\,  \x' \in D, ~|x_D - x'_D| \le \epsilon_D \,\cap\, |x_L - x'_L| \le \epsilon_L \},
\end{equation}
where $\x = (x_D,x_L)$ and $\x' = (x'_D,x'_L)$ indicate the depth and length, and we denote the corresponding stability parameters by $(\epsilon_D,\epsilon_L)$.  This corresponds to $d(\cdot,\cdot)$ being a weighted $\ell_{\infty}$-norm.

We evaluate each algorithm on a $50 \times 50$ grid of points, with the corresponding values coming from the GP posterior that was derived using the original data. We use the Mat\'{e}rn-5/2 ARD kernel, setting its hyperparameters by maximizing the likelihood on a second (smaller) available dataset. 
The parameters $\epsilon_D$ and $\epsilon_L$ are set to $1.0$ and $100.0$, respectively. The stability requirement changes the global maximum and its location, as can be observed in Figure~\ref{fig:lake}. The number of sampling rounds is $T=120$, and each algorithm is initialized with the same $10$ random data points and corresponding observations. The performance is averaged over $100$ different runs, where every run corresponds to a different random initialization. In this experiment, \sgpucb~achieves the smallest $\epsilon$-regret in the early rounds, while in the later rounds \alg~is the only method that reports a near-optimal $\epsilon$-stable point.
\end{document}